\documentclass{article}

\usepackage{graphicx}
\usepackage{booktabs} %
\usepackage{lipsum,mwe,cuted}
\usepackage{float}
\usepackage{caption}
\usepackage{wrapfig}

\usepackage{hyperref}

\usepackage[preprint]{neurips_2025}

\usepackage{amsmath}
\usepackage{graphicx}
\usepackage{amssymb}
\usepackage{mathtools}
\usepackage{amsthm}
\usepackage{xspace}
\usepackage{enumitem}
\usepackage{multirow}
\usepackage{tabularx} %
\usepackage{caption}

\usepackage[utf8]{inputenc} %
\usepackage[T1]{fontenc}    %
\usepackage{url}            %
\usepackage{booktabs}       %
\usepackage{amsfonts}       %
\usepackage{nicefrac}       %
\usepackage{microtype}      %
\usepackage{xcolor}         %
\usepackage{subfig}
\usepackage{wrapfig}

\usepackage[nameinlink,capitalize,noabbrev]{cleveref}

\definecolor{mydarkblue}{rgb}{0,0.08,0.45} %
\hypersetup{
  colorlinks=true,
  frenchlinks=false,
  pdfborder={0 0 0},
  naturalnames=false,
  hypertexnames=false,
  breaklinks,
  linkcolor=mydarkblue,
  citecolor=mydarkblue,
  filecolor=mydarkblue,
  urlcolor=mydarkblue,
}

\theoremstyle{plain}
\newtheorem{theorem}{Theorem}[section]

\theoremstyle{definition}

\theoremstyle{remark}

\newcommand{\method}{DPO-\textit{Shift}\xspace}

\newcommand{\logwin}{\log \pi_\theta(\yw|\x)}
\newcommand{\logrej}{\log \pi_\theta(\yl|\x)}

\newcommand{\pitheta}{\pi_\theta}
\newcommand{\yl}{\boldsymbol{y}_l}
\newcommand{\yw}{\boldsymbol{y}_w}
\newcommand{\x}{\boldsymbol{x}}
\newcommand{\y}{\boldsymbol{y}}
\newcommand{\sx}{\boldsymbol{x}_i}
\newcommand{\syw}{\boldsymbol{y}_{w}^i}
\newcommand{\syl}{\boldsymbol{y}_{l}^i}

\title{DPO-Shift: Shifting the Distribution of Direct Preference Optimization}

\author{%
  Xiliang Yang\thanks{Most of the work of Xiliang Yang was done when he was with School of Data Science, The Chinese University of Hong Kong, Shenzhen.} \\
  School of Mathematics\\
  South China University of Technology \\
  \texttt{xlyangscut@gamil.com} \\
  \And
  Feng Jiang \\
  School of Data Science \\
  The Chinese University of Hong Kong, Shenzhen \\
  \texttt{jiangfeng@cuhk.edu.cn} \\
  \And
  Qianen Zhang \\
  School of Data Science \\
  The Chinese University of Hong Kong, Shenzhen \\
  \texttt{zhangqianen@cuhk.edu.cn} \\
  \And
  Lei Zhao \\
  Institute of Translational Medicine and National Center for Translational Medicine \\
  Shanghai Jiao Tong University \\
  \texttt{zhaolei@sjtu.edu.cn} \\
  \And
  Xiao Li\thanks{Corresponding author.} \\
  School of Data Science \\
  The Chinese University of Hong Kong, Shenzhen \\
  \texttt{lixiao@cuhk.edu.cn} \\
}

\begin{document}

\maketitle

\begin{abstract}
     Direct Preference Optimization (DPO) and its variants have become increasingly popular for aligning language models with human preferences. These methods aim to teach models to better distinguish between chosen (or preferred) and rejected (or dispreferred) responses. However, prior research has identified that the probability of the chosen responses often decreases during training, and this phenomenon is known as likelihood displacement. To tackle this challenge, in this work, we introduce \method to controllably shift the distribution of the chosen probability. Then, we show that DPO-\textit{Shift} exhibits a fundamental trade-off between improving the chosen probability and sacrificing the reward margin, as supported by both theoretical analysis and experimental validation. Furthermore, we demonstrate the superiority of DPO-\textit{Shift} over DPO on a downstream task by a designed win rate experiment. We believe this study shows that the likelihood displacement issue of DPO can be effectively mitigated with a simple, theoretically grounded solution.
\end{abstract}

\section{Introduction}
\label{sec:intro}

There has been a growing interest in guiding large language models (LLMs) to
generate safe and helpful content to align with human values and intentions,
or, taken together, preferences. One of the most important methods in this field is known as Reinforcement Learning from Human Feedback (RLHF) ~\cite{christiano2017deep,
Ouyang2022TrainingLM, stiennon2020learning}. However, multi-stage
optimization procedure is raised in these methods, which includes the training
of a reward model and the policy model to maximize the reward. Such
optimization and computational burden make it challenging to use and analyze,
despite its ability to improve the quality of generated responses ~\cite{bai2022training,
achiam2023gpt, touvron2023llama}.

\textbf{Background and Related Works.} Recently, DPO~\cite{Rafailov2023DirectPO} and its variants \cite{meng2024simpo,azar2024general,tang2024generalized,xu2024contrastive,Ethayarajh2024KTOMA,Park2024DisentanglingLF}
is attracting more and more attention. Given a pair of samples $(\x, \yw, \yl)$ from the dataset, where $\x$ is the prompt, and $\yw$ and $\yl$ represent the chosen and rejected responses, respectively—annotated by strong large language models or humans—the loss of DPO is designed to maximize the margin between the reward of the chosen response and the rejected response for the model $\pitheta$. Being offline algorithms, its simplicity makes DPO more applicable and stable. The main difference between DPO and RLHF lies in the treatment of reward function. DPO proposes to directly parameterize it with the policy model, therefore eliminating the need to train an extra reward model and largely simplifying the training process.

However, it has been reported that both $\logwin$ and $\logrej$ often decrease simultaneously during the training process of DPO; see, e.g., \cite{pal2024smaug, yuan2024advancing, rafailov2024r, tajwar2024preference, pang2024iterative, liu2024provably,razin2024unintentional}. There are several names used to describe such a phenomenon, and we adopt the term ``likelihood displacement'' \cite{razin2024unintentional} in this work. Though DPO still maximizes the reward margin even with this likelihood displacement issue, it remains unfavorable as it causes an unexpected increase in probabilities for responses that are neither preferred nor dispreferred. Unfortunately, this problem hasn't been noticed nor been solved in some of the latest works in this area \cite{xiao2024cal,xiao2025simper,kong2025sdpo,omura2024entropy,pan2025pre,xu2025full}.

Prior work has attributed this phenomenon to limitations in model capacity~\cite{tajwar2024preference}, the presence of multiple training samples or output tokens~\cite{pal2024smaug}, and the initial SFT phase~\cite{rafailov2024r}. Existing studies, such as~\cite{razin2024unintentional}, have provided theoretical insights into addressing this gap and proposed solving the likelihood displacement problem by filtering the datasets.

\textbf{Main Contributions.} In this paper, we propose \method, aiming to solve the likelihood displacement issue of DPO, by adding a parameter function $f(\lambda)$ to the rejected reward in the Bradley–Terry (BT) model \cite{Bradley1952RankAO}, which is detailed in \eqref{eq:lambda_dpo}.

\begin{figure}[H]
\vspace{-10pt}
    \centering
    \subfloat[Distribution comparison of $\log p_{\theta}(\boldsymbol{y} \mid \boldsymbol{x})$ and reward margins using DPO and \method on UltraFeedback (Llama 3-8B). \textbf{Left:} Chosen vs. rejected log-probabilities. \textbf{Right:} Reward margin. \label{fig:teaser}]{
        \includegraphics[width=0.48\linewidth]{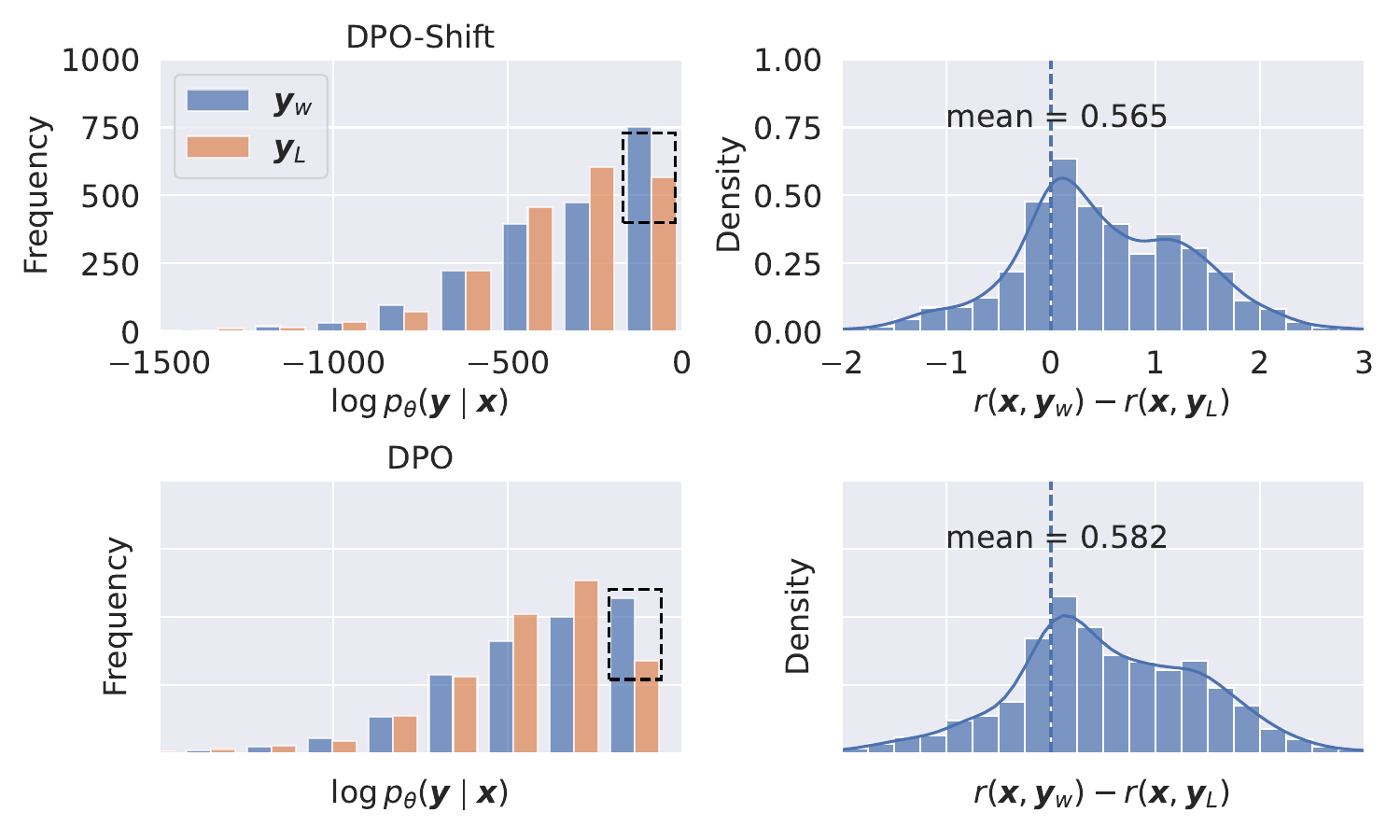}
    }
    \hfill
    \subfloat[Training process for DPO and \method on UltraFeedback (Llama 3-8B) \label{fig:logp_margin_r2}]{
        \includegraphics[width=0.48\linewidth]{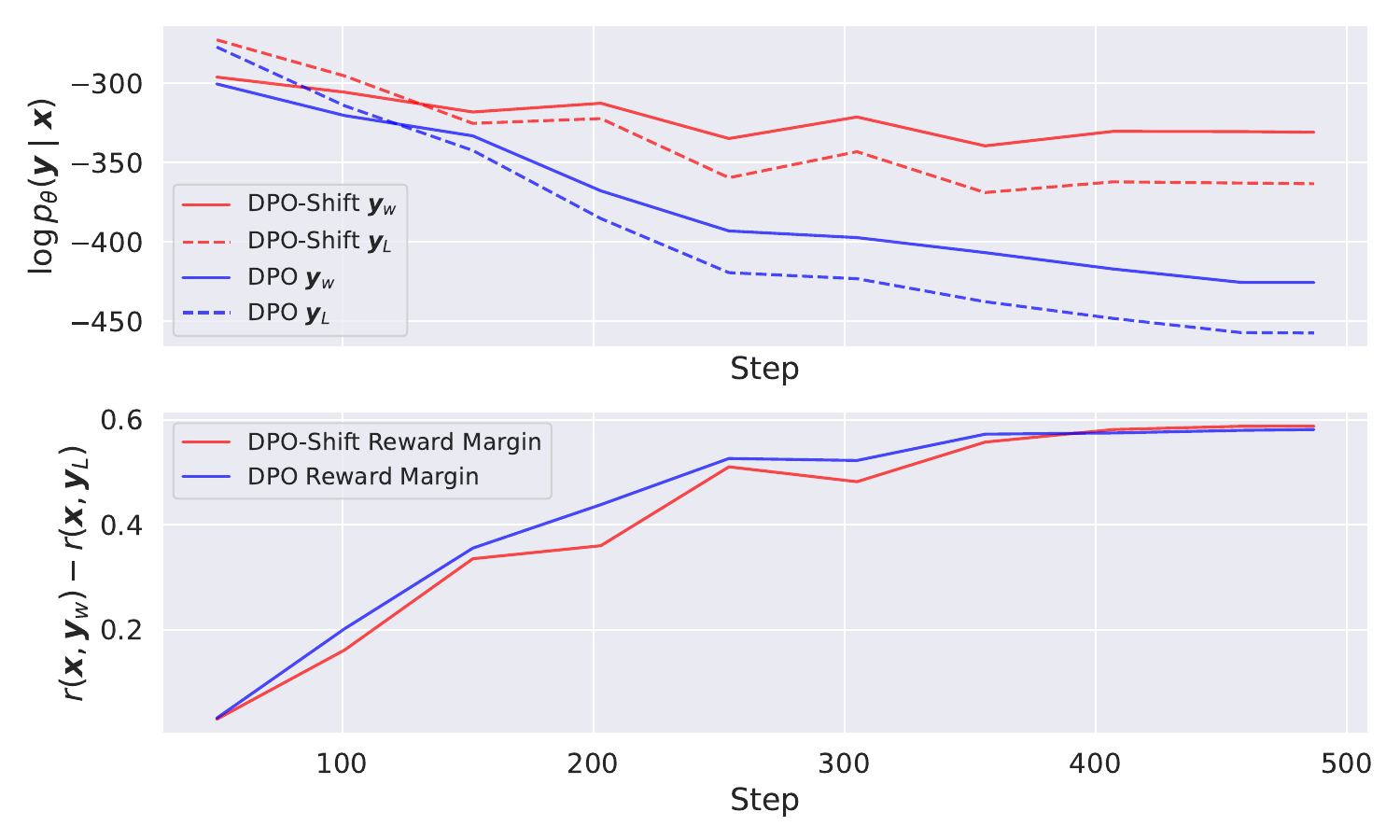}
    }
    \caption{Comparison of log probabilities and reward margins between DPO and \method.}
    \label{fig:logp_margin_comparison_full}
\vspace{-10pt}
\end{figure}

We briefly illustrate in \Cref{fig:teaser} that, by choosing a proper $f(\lambda)$ in \method, we successfully achieve a balance between the distribution of $\logwin$ and the reward margin. The first row corresponds to a specific choice of $f(\lambda)$ of our proposed \method, where we observe an increased chosen probability compared to DPO (depicted in the second row). This improvement is accompanied by only a slight decrease in the reward margin defined as the difference $r(\x,\yw)-r(\x,\yl)$. In fact, we can achieve reward margins that are nearly as high as that of DPO by choosing $f(\lambda)$ properly; see \Cref{sec:exp_results}. The detailed training process of these two methods is established in \Cref{fig:logp_margin_comparison_full}.

Our main contributions can be summarized as follows.

\begin{enumerate}[label=\textup{\textrm{(C.\arabic*)}},topsep=0pt,itemsep=0ex,partopsep=0ex]
    \item We propose \method to mitigate the likelihood displacement issue of DPO by controllably shifting the distribution of the chosen probability. This is achieved through a new parameter function $f(\lambda)$ introduced in \method. Our approach is as simple as DPO and does not require any modifications to the dataset.
    
    \item We provide a theoretical analysis for the proposed \method without imposing additional assumptions. The analysis guarantees that \method mitigates the likelihood displacement issue while introducing a fundamental trade-off. Specifically, our theory reveals that \method improves the chosen probability $\logwin$ at the cost of reducing the reward margin that DPO seeks to maximize. Furthermore, the trade-off is explicitly controlled by $f(\lambda)$, offering practical guidance on its selection. Our findings suggest that $f(\lambda)$ should be chosen relatively close to $1$ to achieve an improvement in the chosen probability over DPO, while only slightly sacrificing the reward margin. 
    
    Experimentally, we conduct thorough ablation studies to train the Llama 3-8B and Qwen 2-7B models on the UltraFeedback and Capybara-preferences datasets with fine grained choices of $f(\lambda)$. The experiment results corroborate our analysis, clearly demonstrating that the likelihood displacement issue is largely mitigated by \method and the fundamental trade-off exists between the chosen probability and reward margin in \method. 

    \item We use downstream experiments to illustrate the improved performance of \method over DPO. In particular, we train the Llama 3-8B and Qwen 2-7B models on the UltraFeedback dataset. Then, to fully demonstrate the superiority that \method can controllably shift the chosen probability, we conduct a designed win rate experiment. The result shows that \method can consistently outperform DPO. 
\end{enumerate}

We believe that our study provides a simple yet efficient approach for mitigating the likelihood displacement issue of DPO.

\section{\method: Formulation and Analysis}
\label{sec:method}

\subsection{DPO, likelihood displacement, and \method}
\label{sec:dpo_pre}

Consider the preference dataset $\mathcal{D}_{\text{pref}}=\{(\x,\yw,\yl)\}$, where $\boldsymbol{x}$ is the prompt, $\boldsymbol{y}_{w}$ and $\boldsymbol
{y}_{l}$ are the chosen and rejected responses
for the prompt $\x$, respectively. Given the prompt $\x$, two responses $\y_1$ and $\y_2$ are first generated from models or
directly taken from static dataset, then the chosen $\yw$ and rejected $\yl$ are selected from $\{\y_1,\y_2\}$ by human or other strong language models. 
DPO consists of
two steps, including supervised fine-tuning (SFT) and preference optimization.

\textbf{Supervised Fine-tuning (SFT).} In this stage, the LLM $\pitheta$ is trained to maximize the log-likelihood of $\y$ given $\x$ with cross-entropy loss:
\begin{align*}
    \min_{\theta} \ \mathcal{L}_{\text{SFT}}(\pitheta) = -\mathbb{E}_{(\boldsymbol{x},\boldsymbol{y})\sim \mathcal{D}_{\text{SFT}}}\left[\log\pitheta(\boldsymbol{y}|\boldsymbol{x})\right],
\end{align*}
Here, $\mathcal{D}_{\text{SFT}} = \{(\x,\y)\}$ is the normal prompt-response dataset used for auto-regressive language modeling and $\pitheta(\y|\x) = \prod_{i=1}^{|y|} \pitheta(y_i|\x, \y_{1:i-1})$. 
For convenience, we refer to the model after the SFT stage as the ``SFTed model".

\textbf{Preference Optimization (PO).} In this stage, DPO
parameterizes the reward model with the LLM $\pi_{\theta}$ via
\begin{align}
    \label{eq:reward:dpo}
    r(\x,\y) = \beta \left(\log \frac{\pi_{\theta}(\y|\x)}{\pi_{\text{ref}}(\y|\x)}+ \log Z(\x)\right),
\end{align}
where $Z(\x)$ is the partition function and $\pi_{\mathrm{ref}}$ is known as the reference model and usually set to be
the ``SFTed model".

Incorporating this reward function into the Bradley-Terry (BT) model, i.e.,
$\mathbb{P}(\yw > \yl|\x) = \sigma(r(\x,\yw)-r(\x,\yl))$. Then, by maximizing the log-likelihood of $\mathbb{P}(\yw > \yl|\x)$, DPO arrives at the following objective function:
\begin{align}
    \label{eq:org_dpo}
    \mathcal{L}_{\text{DPO}}(\pi)=
     -\mathbb{E}\left[\log\sigma\left(\beta\log\frac{\pi_{\theta}(\boldsymbol{y}_{w}|\boldsymbol{x})}{\pi_{\text{ref}}(\boldsymbol{y}_{w}|\boldsymbol{x})}-\beta\log\frac{\pi_{\theta}(\boldsymbol{y}_{l}|\boldsymbol{x})}{\pi_{\text{ref}}(\boldsymbol{y}_{l}|\boldsymbol{x})}\right)\right].
\end{align}

Here, the expectation is taken over the dataset $\mathcal{D}_{\text{pref}}$. The model after the PO state is called the ``POed model".

\textbf{Likelihood Displacement.} Let us recall the distributions of the likelihood $\log \pitheta(\boldsymbol{y}_{w}|\boldsymbol{x})$ and $\log \pitheta(\boldsymbol{y}_{l}|\boldsymbol{x})$ of the DPO model in \Cref{fig:logp_margin_comparison_full}. It is easy to observe that the highest likelihood region for both the chosen and rejected responses are decreased dramatically after DPO. Though some other likelihood regions of the chosen and rejected responses have increased after DPO, the averaged likelihood of $\log \pitheta(\boldsymbol{y}_{w}|\boldsymbol{x})$ and $\log \pitheta(\boldsymbol{y}_{l}|\boldsymbol{x})$ over the entire test set is overall decreasing according to our experiment results, 
aligning with the likelihood displacement phenomenon observed in the existing literature \cite{tajwar2024preference,pal2024smaug,razin2024unintentional,pang2024iterative,yuan2024advancing,liu2024provably,rafailov2024r,Hong2024ORPOMP}.  In conclusion, the likelihood displacement occurs not only in the training stage but also in the test dataset, which is counter-intuitive and can be harmful to the model's generalization ability.

An important factor causing the likelihood displacement issue during the PO stage gives rise to the semantic similarity between the chosen $\yw$ and rejected $\yl$ pairs in $\mathcal{D}_{\text{pref}}$, as observed in the existing works; see, e.g., \cite{tajwar2024preference,Hong2024ORPOMP,razin2024unintentional,pal2024smaug}. This is indeed implied by the generation process of contemporary preference datasets. For instance, the UltraFeedback dataset \cite{Cui2024UltraFeedbackBL} is generated by using different LLMs to response to the same prompt $\x$, and then it selects $\yw$ and $\yl$ using GPT4. To demonstrate it, we pick the following examples from UltraFeedback:
\begin{align*}
    & \text{\parbox{\linewidth} {\text{\textbf{Q1}}: ...Select from female and male... Solution:}}\\
    &\text{\textbf{chosen}: Female.}\\
    &\text{\textbf{rejected}: Female.} \\
    & \text{\parbox{\linewidth} {\text{\textbf{Q2}}: Write the right answer to the question based on...}} \\
    &\text{\textbf{chosen}: Dan, the protagonist, got a coke out of the cooler.}\\
    &\text{\textbf{rejected}: Dan got coke out of the cooler.}
\end{align*}
This partly explains that the model tends to assign
similar probabilities to both responses. In the DPO objective function, it seeks to maximize the margin between the
probability of the chosen and rejected responses even if they are semantically similar. Hence, instead of the ideal case where it maximizes the chosen probability while minimizes the rejected one, it often reduces the probability of both
responses with similar semantic structures, though their margin is enlarged. This leads to the likelihood displacement issue. Consequently, the model favors
responses that are neither chosen nor rejected.

\textbf{\method.} To address this counter-intuitive and harmful likelihood displacement issue of DPO, we introduce \method in this work. The motivation behind the proposed method is to alleviate the problem caused by the similarity of the chosen and rejected pairs. As we analyzed previously, the chosen probability decreases accordingly when the DPO objective maximizes the margin between two semantically similar responses. Based on this observation, 
we propose to add a real-valued function $0<f(\lambda)<1$ to
the reward of the rejected response. This helps the BT model to rank correctly by reducing the confrontation between two semantically similar responses, potentially mitigating the likelihood displacement issue of DPO. Mathematically,  our proposed formulation is displayed in the following:
\begin{align}
    \mathcal{L}_{\text{DPO-Shift}}(\pi) =  -\mathbb{E}&\left[\log\sigma \left(\beta\log\frac{\pi_{\theta}(\boldsymbol{y}_{w}|\boldsymbol{x})}{\pi_{\mathrm{ref}}(\boldsymbol{y}_{w}|\boldsymbol{x})}  -f(\lambda)\cdot\beta\log\frac{\pi_{\theta}(\boldsymbol{y}_{l}|\boldsymbol{x})}{\pi_{\mathrm{ref}}(\boldsymbol{y}_{l}|\boldsymbol{x})}\right)\right].\label{eq:lambda_dpo}
\end{align}

\subsection{Analysis for \method}
\label{sec:theory}

We analyze the effects of \method for two important quantities, including the likelihood of the chosen response $\log \pi_{\theta}(\yw|\x)$ and the indicator function of the reward margin $\mathbf{1}\{(\x,\yw,\yl)|\log \frac{\pi_{\theta}(\yw|\x)}{\pi_{\text{ref}}(\yw|\x)}-\log \frac{\pi_{\theta}(\yl|\x)}{\pi_{\text{ref}}(\yl|\x)} >0\}$. The latter reflects the model's ability to align with human preferences and is implicitly maximized by DPO's objective. We define the two target functions as follows:
\begin{align}
    \omega_{1}(\theta) & =\mathbb{E}\left[\log\pi_{\theta}\left(\boldsymbol{y}_{w}|\boldsymbol{x}\right)\right],\label{eq:ob_cho}      \\
    \omega_{2}(\theta) & =\mathbb{E}\left[\mathbf{1}\left\{\log\frac{\pi_{\theta}(\boldsymbol{y}_{w}|\boldsymbol{x})}{\pi_{\text{ref}}(\boldsymbol{y}_{w}|\boldsymbol{x})} -\log\frac{\pi_{\theta}(\boldsymbol{y}_{l}|\boldsymbol{x})}{\pi_{\text{ref}}(\boldsymbol{y}_{l}|\boldsymbol{x})}>0\right\}\right].\label{eq:ob_acc}
\end{align}
The likelihood displacement issue of DPO enlarges $\omega_2$ while decreases $\omega_1$. To provide an analytical characterization, we alter the discontinuous $\omega_{2}$ and consider its smoothed version
\begin{align}
    \label{eq:ob_smoo_acc}\omega_{2}(\theta)=\mathbb{E}\left[\sigma\left(\gamma\log\frac{\pi_{\theta}(\boldsymbol{y}_{w}|\boldsymbol{x})}{\pi_{\text{ref}}(\boldsymbol{y}_{w}|\boldsymbol{x})} -\gamma\log\frac{\pi_{\theta}(\boldsymbol{y}_{l}|\boldsymbol{x})}{\pi_{\text{ref}}(\boldsymbol{y}_{l}|\boldsymbol{x})}\right)\right],
\end{align}
where $\gamma$ is the smoothing factor. To compare \method with the original DPO, we introduce two functions measuring the gaps between targets after one step of optimization (i.e., updating $\theta_t$ to $\theta_{t+1}$) with different objective functions:
\begin{align}
    g_{1}(t+1) = \omega_{1}(\theta_{t+1})\Big|_{\text{DPO-Shift}}-\omega_{1}(\theta_{t+1})\Big|_{\text{DPO}},
    g_{2}(t+1) = \omega_{2}(\theta_{t+1})\Big|_{\text{DPO-Shift}}-\omega_{2}(\theta_{t+1})\Big|_{\text{DPO}}.
\end{align}
We characterize the two gap functions
in the following theorem.
\begin{theorem}
    \label{thm:gap} Given $\theta_{t}$ and learning rate $\eta$ and denote
    \begin{align*}
        c(\theta)        & =\gamma\sigma\left(f(\lambda)\cdot\gamma\log\frac{\pi_{\theta}(\boldsymbol{y}_{l}|\boldsymbol{x})}{\pi_{\mathrm{ref}}(\boldsymbol{y}_{l}|\boldsymbol{x})}-\gamma\log\frac{\pi_{\theta}(\boldsymbol{y}_{w}|\boldsymbol{x})}{\pi_{\mathrm{ref}}(\boldsymbol{y}_{w}|\boldsymbol{x})}\right), \\
        \eta_1(\theta) & =\eta \sigma \left(\log\frac{\pi \left(\boldsymbol{y}_{l}|\boldsymbol{x}\right)}{\pi_{\mathrm{ref}}{(\boldsymbol{y}_{l}|\boldsymbol{x})}}-\log \frac{\pi \left(\boldsymbol{y}_{w}|\boldsymbol{x}\right)}{\pi_{\mathrm{ref}}(\boldsymbol{y_{w}|x})}\right).
    \end{align*}
    We have
     \begin{align}\label{eq:gap characterization}
     g_{1}(t+1) = (1-f(\lambda)) u_1, \quad g_{2}(t+1) = (1-f(\lambda)) u_2.
    \end{align}
    Here, 
    \begin{equation}\label{eq:u factors} 
    \begin{aligned}
       &u_1 = \mathbb{E}\left[c(\theta)\cdot\nabla_{\theta}\log\pi_{\theta}\left(\boldsymbol{y}_{l}|\boldsymbol{x}\right)^{\top}\nabla_{\theta}\log\pi_{\theta}\left(\boldsymbol{y}_{w}|\boldsymbol{x}\right)\right], \\
       & u_2 = \mathbb{E}\left[\eta_{1}(\theta)\left(\nabla_{\theta}\log\pi_{\theta}\left(\boldsymbol{y}_{l}|\boldsymbol{x}\right)^{\top}\nabla_{\theta}\log\pi_{\theta}\left(\boldsymbol{y}_{w}|\boldsymbol{x}\right) - \left\|\nabla_{\theta}\log\pi_{\theta}\left(\boldsymbol{y}_{l}|\boldsymbol{x}\right)\right\|^{2}\right)\right].
    \end{aligned}
    \end{equation}
\end{theorem}
The proof of Theorem 2.1 is deferred to \Cref{app:proof}. It is worth mentioning that our derivation in the proof of \Cref{thm:gap} applies to every single sample, and hence the result in \Cref{thm:gap} applies to $u_1$ and $u_2$ defined using any specific dataset. We state the results in expectation in order to be data-independent.

This theorem explains the pros and cons of \method, yielding indications for choosing $f(\lambda)$. We provide detailed discussions below.

\textbf{Fundamental Trade-off.} We are interested in characterizing the sign of the two gap functions using \eqref{eq:gap characterization}. To compute $u_1$ and $u_2$ on a specific $\mathcal{D}_{\text{pref}}$, we define the following sample-based version:
\begin{align*}
    u^{i}_{1} & = c_{i}(\theta)\cdot\nabla_{\theta}\log\pi_\theta \left(\syl|\sx\right)^{\top}\nabla_{\theta}\log\pi_\theta \left(\syw|\sx\right) ,            \\
    u^{i}_{2} & = \eta_{1}\left(\nabla_{\theta}\log\pi_\theta \left(\syl|\sx\right)^{\top}\nabla_{\theta}\log\pi_\theta \left(\syw|\sx\right) - \left\|\nabla_{\theta}\log\pi_\theta \left(\syl|\sx\right)\right\|^{2}\right).
\end{align*}
Then, we compute the sample average $u_1 = \sum_{i}u^{i}_{1} / |\mathcal{D}_{\text{ref}}|$ and $u_2 = \sum_{i}u^{i}_{2} / |\mathcal{D}_{\text{ref}}|$. On the test set of UltraFeedback, when setting $\gamma=1$ and 
$\pi_\theta$ to be the SFTed Llama 3-8B, we obtain that $u_1 = 4.84\times 10^8$. For $u_2$, since $\eta_1$ is bounded, we set it to be 1 and obtain $u_2 = -4.28\times 10^9$. In terms of frequency, $71.4\%$ of $\{u_1^i\}$ are positive and $81.7\%$ of $\{u_2^i\}$ are negative. These results indicate a clear positivity of $u_1$ while a clear negativity of $u_2$. Indeed, positivity of $u_1$ is expected due to the semantic similarity between $\yw$ and $\yl$ in contemporary $\mathcal{D}_{\text{pref}}$.

Since we choose $0<f(\lambda) < 1$, we have $1-f(\lambda) >0$. In this case, $g_1 >0$ as $u_1>0$. This immediately concludes that \method improve the chosen probability (or likelihood) $\log \pi_{\theta}(\yw|\x)$ compared to DPO, solving the undesired likelihood displacement issue. However, there is an explicit trade-off to achieving this improvement. Since $u_2$ can be negative as shown in our test result, $g_2$ can be negative, leading to reduced reward margin of \method compared to DPO. 

In summary, \method improves the chosen probability over DPO, at the cost of decreasing the reward margin. This also yields the indications for choosing $f(\lambda)$, which we describe below.

\textbf{Indications for Choosing $f(\lambda)$.} As analyzed previously, the important trade-off exists in \method. A slightly deeper analysis reveals that a smaller $f(\lambda)$ leads to more increase in chosen probability while a more severe drop in the reward margin. Thus, this indicates that choosing a relatively large $f(\lambda)$, i.e., close to $1$, helps to balance both sides. That is, the chosen probability is improved reasonably compared to DPO, in the meanwhile the reward margin of \method is only decreased slightly. The balance controlled by $f(\lambda)$ is thoroughly demonstrated by our experiment results in \Cref{sec:exp_results}.

For the strategy of choosing of $f(\lambda)$, the first one is to fix it all along the optimization process, i.e., $f(\lambda) = \lambda$. We denote this strategy as \texttt{fixed}. We also propose to vary $\lambda$ between the minimal $\lambda_{\min}$ and the maximal $\lambda_{\min}$ along with time $t$. We denote this strategy as $f(\lambda_{\min},\lambda_{\max},t)$. In this paper, we mainly have linear increase/decrease between $\lambda_{\min}<1$ and $\lambda_{\max}=1$. Set the maximal iteration steps to be $T$, the detailed strategy for the linear increase version of $f(\lambda_{\min},\lambda_{\max},t)$ is $t/T(\lambda_{\max}-\lambda_{\min})+\lambda_{\min}$, while the linearly decreased version is $t/T(\lambda_{\min}-\lambda_{\max})+\lambda_{\max}$. They are separately denoted as \texttt{linear\_increase} and \texttt{linear\_decrease}.

\textbf{Can $f(\lambda)$ be Chosen Larger than 1?} By default, we choose $f(\lambda) <1$ in \method to achieve chosen probability improvement, which is based on the hypothesis that $u_1$ is generally positive. If we encounter the case where $u_1<0$, e.g.,  when most pairs $\yw$ and $\yl$ are dissimilar to each other, $u_2$ must be negative as well. Interestingly, in this case \Cref{thm:gap} suggests that \method with $f(\lambda) >1$ can lead to simultaneous improvements for both the chosen probability and reward margin. However, the event $u_1<0$ is likely to be very rare, given the general similarity between $\yw$ and $\yl$ in existing $\mathcal{D}_{\text{pref}}$. Using $f(\lambda) >1$ when $u_1>0$ leads to catastrophic results. For instance, we trained Llama 3-8B on UltraFeedback with fixed $f(\lambda) = 1.05$, and the model quickly exhibited crash behavior, producing unreadable responses. Therefore, choosing $f(\lambda) > 1$ should be done with great care unless $u_1$ is clearly negative and the overwhelming majority of $\{u_1^i\}$ has negative values. 

Though choosing $f(\lambda) > 1$ is highly discouraged, the above analysis provides a possible direction to further improve \method. Before implementing the PO state, we can first calculate $\{u_1^i\}$ for the entire training set. Then, we assign a specific $f(\lambda_i)$ for each data point, i.e., $f(\lambda_i)<1$ when $u_1^i > 0$ and $f(\lambda_i)>1$ when $u_1^i <0$. Adopting such a carefully crafted approach requires a significant amount of additional implementation code and falls beyond the scope of this work. We leave it as future work.

\section{Experimental Results}
\label{sec:exp_results}

In this section, we present the main results of our experiments, highlighting the performance of \method on various benchmarks and ablation studies.

\subsection{Experimental Setup}
\label{sec:exp_setup}

We perform our experiment on two models, Llama 3-8B~\cite{llama3modelcard}
and Qwen 2-7B~\cite{qwen2}, under base setup. We mainly carry out the following two parts of experiments.

\textbf{Verification Experiment.} This part of the experiment is designed to validate the theoretical results presented in \Cref{sec:theory}. Specifically, we consider three strategies for selecting $f(\lambda)$ in \eqref{eq:lambda_dpo}: \texttt{fixed}, \texttt{linear\_increase}, and \texttt{linear\_decrease}. We design a series of ablation studies for each of the strategy by altering the choice of $f(\lambda)$. For the \texttt{fixed} strategy, we perform an ablation study by evaluating fixed $f(\lambda)$ from the range $[0.5, 0.55, 0.6, 0.65, 0.75, 0.8, 0.85, 0.9, 0.95]$. For the \texttt{linear\_increase} and \texttt{linear\_decrease} strategies, we set $\lambda_{\min}$ to values in $[0.75, 0.85, 0.95]$ and fix $\lambda_{\max} = 1$. We compute the $\logwin$, $\logrej$, and reward margin for models trained with these $f(\lambda)$ strategies on the test sets of their respective training datasets. To further illustrate the phase transition phenomenon in these probability distributions as $f(\lambda)$ varies from $f(\lambda) < 1$ to $f(\lambda) = 1$, we extend the \texttt{fixed} strategy ablation study by including $f(\lambda)$ values of $[0.96, 0.97, 0.98, 0.99]$.

\textbf{Downstream Performance Evaluation.} This experiment is primarily designed to evaluate the general performance of the model trained using our proposed method. For the two SFTed models trained on the UltraFeedback dataset, we evaluate the win rate of \method against DPO using 2,000 randomly selected questions from the Capybara dataset and the \href{https://huggingface.co/datasets/HuggingFaceH4/ultrafeedback_binarized/viewer/default/test_prefs}{\texttt{test\_prefs}} split of the UltraFeedback dataset. For the evaluation, we employ the \href{https://ollama.com/library/llama3.3:70b-instruct-q4_K_M}{Llama3.3:70b-instruct-q4\_K\_M} model provided in the ollama library, a 4-bit quantized version of the latest Llama 3.3-Instruct 70B, which delivers performance comparable to Llama 3.1-405B.

Additional experiment details are deferred to \Cref{app:exp_details_app}.

\subsection{Experimental Verification for \method} \label{sec:chosen_shift}

We report some representative results from our ablation studies for \texttt{fixed}, \texttt{linear\_increase} and \texttt{linear\_decrease}. We mainly evaluate them with 2 metrics, including distribution for $\logwin$ and $\logrej$, and distribution for reward margin. 

\begin{figure}[t]
    \centering
    \includegraphics[width=\linewidth]{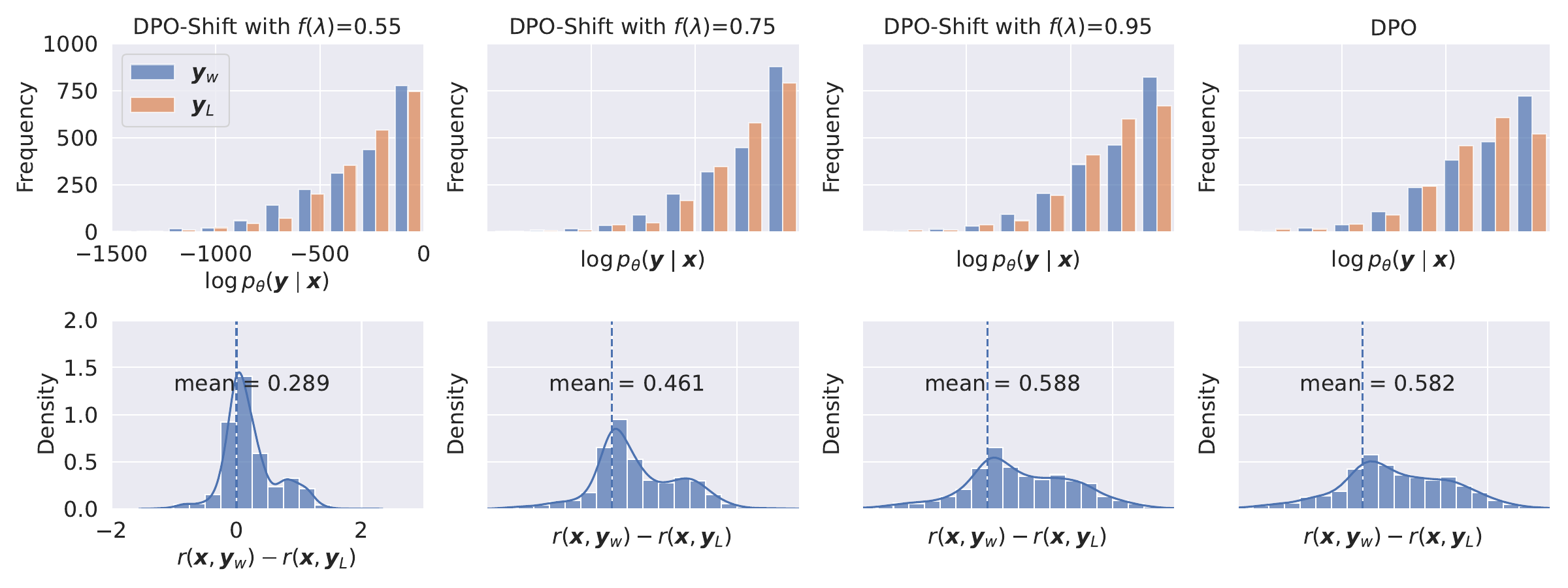}
    \caption{Distribution for $\logwin$, $\logrej$ (up) and reward margin as well as the mean on test set split of UltraFeedback for Llama 3-8B trained on UltraFeedback with \texttt{fixed} strategy. Only limited cases of $f(\lambda)$ are listed. For a full ablation study, please refer to \Cref{app:candr_supp}. The ranges of the y-axis of all subfigures are the same.}
    \label{fig:ll_all_fix_ultra}
    \vspace{-0.5cm}
\end{figure}

We first study \texttt{fixed} strategy. The two metrics for evaluation, including distributions for $\logwin$ and $\logrej$, and distribution for reward margin are demonstrated in \Cref{fig:ll_all_fix_ultra}. Thorough ablation experiment results on different datasets and models are supplemented in \Cref{app:candr_supp}, where we have more choices of $f(\lambda)$ for Llama 3-8B trained on UltraFeedback, new experiments for Llama 3-8B trained on Capybara, new experiments for Qwen 2-7B trained on UltraFeedback, and new experiments for Qwen 2-7B trained on Capybara. In \Cref{fig:ll_all_fix_ultra}, we observe a consistent ``shift'' in the distribution of $\logwin$ and $\logrej$, moving from high-probability regions to lower-probability regions with increasing $f(\lambda)$. This shift is accompanied by a mass transition in the reward margin, shifting from areas less than 0 to areas greater than 0. The extent of this ``shift'' largely depends on the value of $f(\lambda)$.
Specifically, when $f(\lambda)$ is very small, it leads to higher probabilities for the language model on the chosen answer $\boldsymbol{y}_w$. However, this comes at the cost of decreased and shifted reward margin. Thus, small $f(\lambda)$ can result in ``over-fitting" to the chosen answer, losing the contrastive learning ability against the rejected one and reducing the model's performance. Fortunately, by selecting a relatively larger $f(\lambda)$ that is closer to 1, an increase in the reward margin with only a limited trade-off in the chosen probability is observed. This aligns with our analysis proposed in \Cref{sec:theory} that larger $f(\lambda)$ helps balance both metrics. 
For example,  $f(\lambda)=0.9, 0.95$ achieves a higher probability of the chosen responses than DPO while maintaining almost the same reward margin. 

\begin{figure}[t]
\centering
\begin{minipage}[t]{0.48\textwidth}
  \centering
  \includegraphics[width=\linewidth]{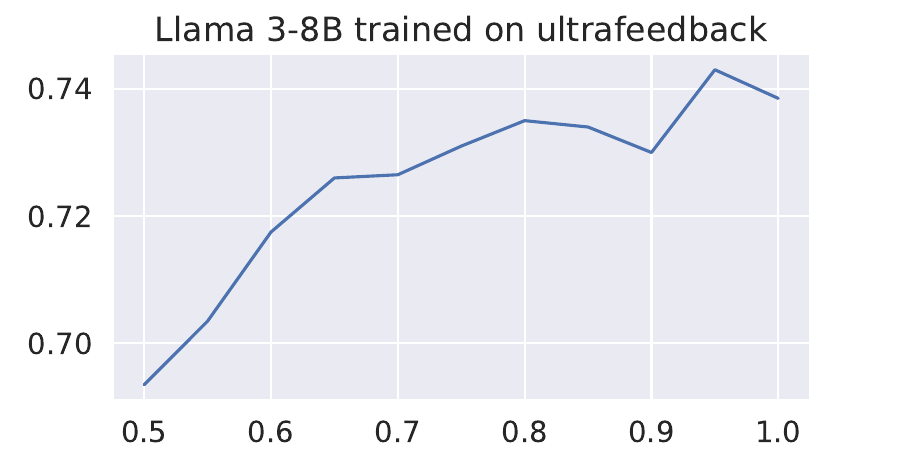}
  \vspace{-0.2cm}
  \caption{Reward accuracy vs different $f(\lambda)$ on the UltraFeedback test set for Llama 3-8B trained on UltraFeedback with \texttt{fixed}, where $f(\lambda)$ is selected from 0.5 to 0.95.}
  \label{fig:ll_rew_vs_acc_fix_ultra}
\end{minipage}%
\hfill
\begin{minipage}[t]{0.48\textwidth}
  \centering
  \includegraphics[width=\linewidth]{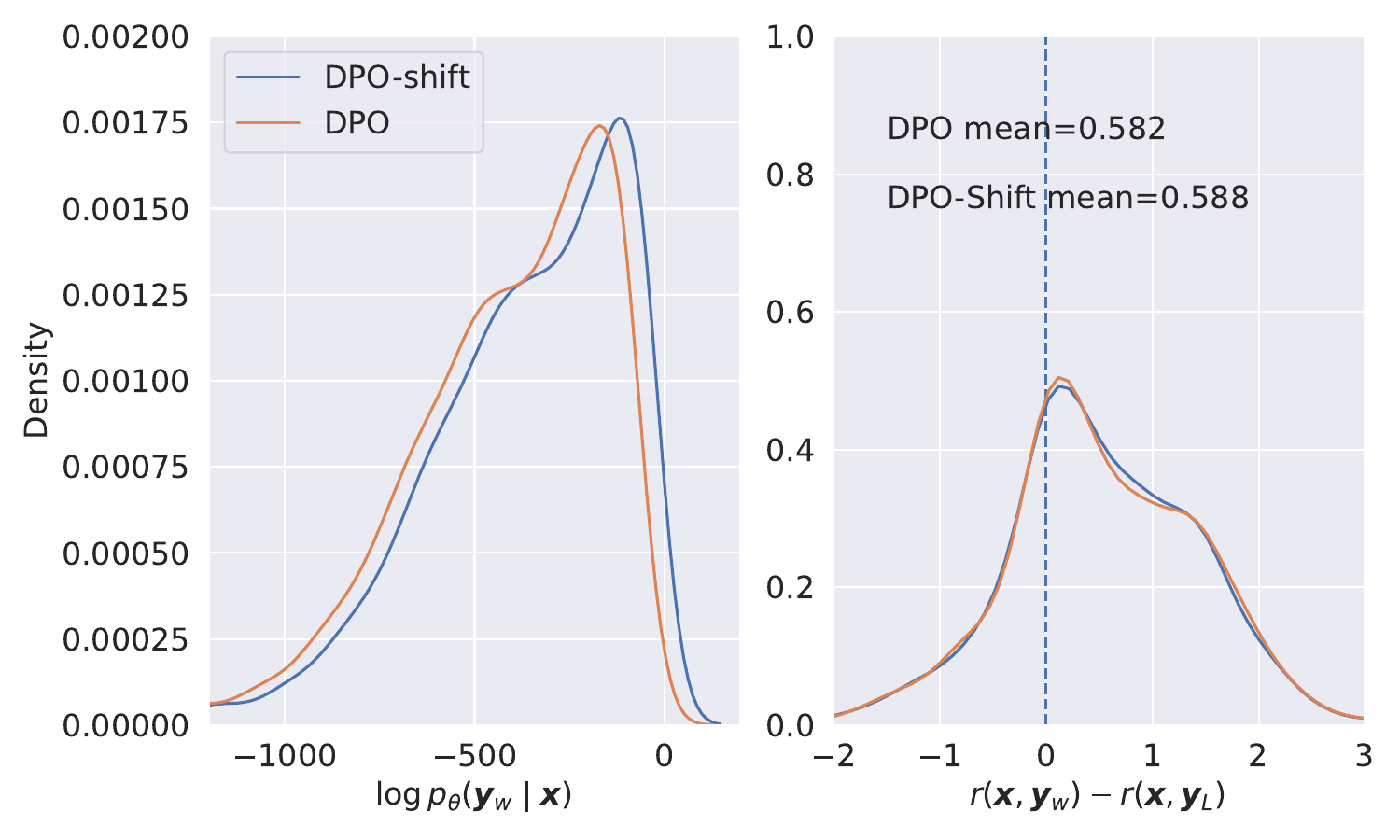}
  \vspace{-0.2cm}
  \caption{The comparison between \method{} with $f(\lambda)=0.99$ and DPO on the UltraFeedback test set for Llama 3-8B. \textbf{Left:} Distribution for $\logwin$. \textbf{Right:} Reward accuracy and margin distribution.}
  \label{fig:ll_rew_fix_ultra}
\end{minipage}
\vspace{-0.5cm}
\end{figure}

In addition to the reward margin, we also display the reward of \method with several choices of $f(\lambda)$ and DPO in \Cref{fig:ll_rew_vs_acc_fix_ultra}, where the reward accuracy is defined as the frequency ratio of $r(\x,\yw)-r(\x,\yl)>0$ over the test dataset split. Similar to the reward margin, it illustrates a clear increase in reward accuracy along with increasing $f(\lambda)$, further corroborating our theory.

\begin{figure}[t]
    \centering
    \includegraphics[width=0.9\textwidth]{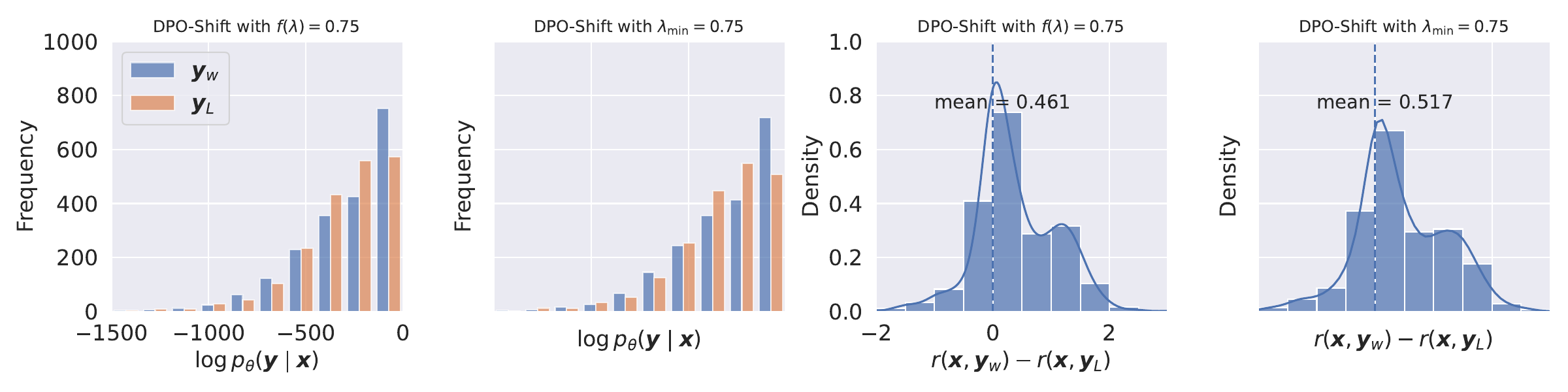}
    \vspace{-0.2cm}
    \caption{Comparison between \texttt{fixed} strategy with $f(\lambda)=0.75$ and \texttt{linear\_decrease} with $\lambda_{\min}=0.75$ on the test set split of UltraFeedback for Llama 3-8B trained on UltraFeedback. \textbf{Left:} Distribution for $\logwin$ and $\logrej$. \textbf{Right:} Reward accuracy and distribution for reward margin.}
    \label{fig:ll_rew_acc_candr_var}
\end{figure}

Additionally, we conduct experiments with more fine grains to investigate the effect of $f(\lambda)$ on the distribution of $\logwin$ and $\logrej$ as it approaches 1. In the case of \texttt{fixed}, we select a new set of fixed $f(\lambda)$s, including $[0.96,0.97,0.98,0.99]$. We report the comparison between \texttt{fixed} with $f(\lambda)=0.99$ and the original DPO with the distribution of chosen probability and reward margin in the \Cref{fig:ll_rew_fix_ultra}. Compared to DPO, an obvious shift of the distribution $\logwin$ can be observed, while the distribution of the reward margin remains nearly unchanged. The full experimental results for $f(\lambda) \in [0.96,0.97,0.98,0.99]$ are reported in \Cref{app:phase_compare}, in which the results are consistent with our analysis.

To achieve a possibly better trade-off between the two distributions, we conduct experiments on \texttt{linear\_increase} and \texttt{linear\_decrease} strategies for choosing $f(\lambda)$. We report the result for \texttt{linear\_decrease} with $\lambda_{\min} = 0.75 $ in \Cref{fig:ll_rew_acc_candr_var}. It can be seen that this dynamic $f(\lambda)$ strategy achieves a more satisfactory balance between reward accuracy and $\logwin$ compared to \texttt{fixed} with $f(\lambda) = 0.75$. Thorough ablation experiment results for \texttt{linear\_increase} and \texttt{linear\_decrease} strategies are provided in \Cref{app:candr_var_supp}, where we include more fine-grained choices of $\lambda_{\min}$, the dataset Capybara, and the model Qwen 2-7B.

\textbf{Summary.} By setting $f(\lambda)$ properly in \method, we can achieve controllable trade-off between chosen probability and reward accuracy. A carefully chosen For example, if one chooses $f(\lambda) = 0.95$ in \method, the likelihood displacement issue can be largely mitigated, while the reward margin can be kept nearly unchanged compared to DPO.

\subsection{Downstream Performance} \label{sec:win_rate_test}

When comparing the answers generated by DPO and \method, we observe that the model trained with DPO generally generates longer and less meaningful responses, while \method tends to produce more concise and high-quality responses. 
Based on these observations, we adopt perplexity, a widely used evaluation metric \cite{cooper2024perplexed,chen1998evaluation,xu2022systematic} for language models. Perplexity quantifies how well a probability model predicts a sample of data, essentially measuring the model's uncertainty about the data. Lower perplexity indicates that the model is better at predicting the sample. The perplexity of DPO and \method trained on UltraFeedback with \texttt{fixed} $f(\lambda) = 0.95$ is evaluated on the chosen responses from the test split of the UltraFeedback dataset. The results are 4.475 for \method and 18.996 for DPO, further demonstrating the potential advantage of \method. 

\begin{table}[H]
\centering
\begin{tabular}{lcccc}
\toprule
$f(\lambda)$ strategy &  \textbf{Win} & \textbf{Lose} \\
\midrule
\textbf{SFT} & 34.20\% & 65.80\% \\
\midrule
\texttt{fixed} 0.55 & 51.40\% & 49.60\%  \\
\texttt{fixed} 0.75 & 57.30\% & 42.70\%  \\
\texttt{fixed} 0.95 & 68.95\% & 31.05\%  \\
\texttt{linear\_increase} 0.95 & 67.40\% & 32.60\% \\
\texttt{linear\_decrease} 0.95 & 72.15\% & 27.85\% \\
{SimPO} \cite{meng2024simpo} & 56.75\% & 43.25\% \\
{KTO} \cite{ethayarajh2024kto} & 61.00\% & 39.00\% \\
{IPO} \cite{azar2024general} & 60.50\% & 39.50\% \\
\bottomrule
\end{tabular}
\caption{Win rate experiment against DPO using Llama 3-8B trained on the UltraFeedback dataset and tested with questions from the test split of UltraFeedback. Results for \method using all three strategies including \texttt{fixed}, \texttt{linear\_increase}, \texttt{linear\_decrease} are shown.}
\label{table:win_rate}
\end{table}

To fully demonstrate the better alignment of \method with the chosen response, we conduct a win rate experiment. In this setup, the judge model is provided with the question, the reference answer from the dataset, and the answers generated by \method and DPO. The judge model then judges the responses of \method and DPO based on their general quality and how close to the reference answer they are. We put the judge prompts in \Cref{app:prompts}.

We test the win rate using the test split of UltraFeedback for Llama 3-8B, which is trained on UltraFeedback. In \Cref{table:win_rate}, we report the results for \method and other preference optimization methods including SimPO \cite{meng2024simpo}, KTO \cite{ethayarajh2024kto} and IPO \cite{azar2024general}. To ensure every method is tested at its best performance, we directly adopt the checkpoints provided by SimPO\footnote{\url{https://github.com/princeton-nlp/SimPO?tab=readme-ov-file}}, where extensive hyperparameter grid search has been conducted. Note that the experimental setups for the SimPO, KTO, and IPO checkpoints are identical to ours, as we strictly follow the same configuration as SimPO. Therefore, these models are directly comparable. 

One can observe from \Cref{table:win_rate} that \method consistently outperforms DPO once $f(\lambda)$ is chosen properly, i.e., when $f(\lambda)$ is relatively closer to $1$, corroborating our analysis in \Cref{sec:theory} and the verification experiments in \Cref{sec:chosen_shift}. Specifically, \method wins DPO $\frac{72.15\% - 27.85\%}{2} = 22.15 \%$ more if the \texttt{linear\_decrease} with $\lambda_{\min} = 0.95$ strategy is chosen for \method. Our method also exhibits superior performance compared to SimPO, KTO, and IPO, as \method wins DPO more compared to these methods. For instance, KTO wins $61\%$ against DPO, while \method using \texttt{linear\_decrease} with $\lambda_{\min} = 0.95$ wins $72.15\%$ against DPO, further illustrating the advantage of our method.  In \Cref{app:full: win rate}, we also display the win rate experiment of the Qwen 2-7B model. As SimPO does not provide checkpoints for Qwen 2-7B, we only compare \method against DPO, while omitting other methods. The results for Qwen 2-7B align with our observations from \Cref{table:win_rate}

\textbf{Summary.} \method outperforms DPO in terms of downstream performance when $f(\lambda)$ is set properly to achieve a good balance between the chosen probability and the reward margin.

\subsection{Ablation Study: Comparison with Other Potential Methods}

To clearly distinguish \method from other baseline alternatives, we supplement the comparison on the mean of $\logwin$, $\logrej$, and the reward margin results with other existing preference optimization methods including IPO, SimPO, and KTO. To lift $\logwin$ in the DPO stage, we note that it is common practice in the community to combine the original DPO objective with the SFT loss, which is denoted as $\alpha$-DPO in our paper:
\begin{align*}
    \mathcal{L}_{\text{DPO-}\alpha}(\pi)=\mathcal{L}_{\text{DPO}}(\pi)-\frac{\alpha}{|\yw|}\log \pi(\yw|\x),
\end{align*}
This approach may also mitigate the likelihood displacement issue by properly selecting $\alpha$. The experiment results are presented in \Cref{tab:performance_metrics}, where \method uses $f(\lambda) = 0.95$.

We tried different $\alpha$ and found that $\alpha=0.1$ gives almost the same log probability of the chosen response compared to \method. It can be observed that $\alpha$-DPO works, as it mitigates the likelihood displacement issue. However, though it achieves almost the same log probability of the chosen response as \method, it has a higher log probability of the rejected response. This is reasonable as $\alpha$-DPO lifts the probability of $\yw$ by directly adding a SFT term on $\yw$, at the price of reducing the effects of the DPO term and hence weakening the confrontation between $\yw$ and $\yl$. Consequently, the reward margin of this approach decreased and is inferior to \method. As for other existing preference optimization methods, \method demonstrates consistent advantages on all metrics compared to SimPO and KTO. In the case of IPO, although it exhibits a higher reward margin, it does so at the cost of significantly lower chosen log probability, which further worsens the likelihood displacement issue. 

\textbf{Summary.} The fundamental property of \method lies in that it can increase $\logwin$ (i.e., mitigating the likelihood displacement issue), while keeping the reward margin nearly unchanged compared to DPO once $f(\lambda)$ is chosen properly. This, together with \Cref{table:win_rate}, further justifies the superiority of \method.

\begin{table}[h]
\centering
\begin{tabular}{lccccccc}
\toprule
 Metric & SFT & DPO-Shift & DPO & IPO & SimPO & KTO & $\alpha$-DPO \\
 \midrule
$\logwin$ & -299.09 & -330.86 & -425.47 & -591.85 & -404.46 & -360.93 & -314.59  \\ 
 $\logrej$ & -278.60 & -363.67 & -457.25 & -644.37 & -412.29 & -420.29 & -343.73\\ 
Reward margin & N/A & 0.58 & 0.58 & 0.73 & 0.28 & 0.49 & 0.49\\ 
\bottomrule
\end{tabular}
\vspace{0.2cm}
\caption{Comparison between \method and other potential methods.}
\label{tab:performance_metrics}
\vspace{-0.3cm}
\end{table}

\section{Conclusion and Discussions on Limitations}
\label{sec:conclusion}
In this work, we proposed \method, which controllably shifts the distribution of the chosen probability. Our method guarantees to mitigate the likelihood displacement issue of DPO while introducing a fundamental trade-off between the chosen probability and reward margin. By selecting $f(\lambda)$ carefully, the chosen probability can be improved, in the meanwhile the reward margin is only slightly sacrificed. Extensive ablation experiments across various models and datasets confirm the validity of our theoretical findings. To further demonstrate the advantages of \method, we conducted experiments on downstream task using a designed win rate experiment. Clear performance improvements over DPO were observed, highlighting the robustness and effectiveness of our approach. 

\textbf{Limitations.} We only consider a global $f(\lambda)$ for all the data points. A more crafted strategy on the selection of $f(\lambda)$ can be a possible direction to further improve \method, as we commented at the end of \Cref{sec:theory}. We leave it as future work.

\bibliographystyle{plain} %
\bibliography{ref}  %

\clearpage
\tableofcontents

\clearpage
\appendix

\section{Supplemented Experimental results}
\label{app}

\subsection{Additional Experimental details}
\label{app:exp_details_app}

For the \textbf{SFT} stage, we train our models for one epoch on \href{https://huggingface.co/datasets/HuggingFaceH4/ultrachat_200k}{UltraChat-200k} dataset~\cite{Ding2023EnhancingCL} to obtain an SFT model with a learning rate of 2e-5. For Llama 3-8B model, we directly use the off-the-shelf models from \citep{meng2024simpo} (\href{https://huggingface.co/princeton-nlp/Llama-3-Base-8B-SFT}{princeton-nlp/Llama-3-Base-8B-SFT}), which follows the same training strategy.

For the \textbf{Preference optimization} stage, we aim to verify the robustness of \method. To this end, we perform preference
optimization on \href{https://huggingface.co/datasets/HuggingFaceH4/ultrafeedback_binarized}{UltraFeedback}
dataset~\cite{Cui2024UltraFeedbackBL},
\href{https://huggingface.co/datasets/argilla/Capybara-Preferences}{Capybara-preferences}
dataset~\cite{cbdatacard}. Notice that since the
Capybara dataset lacks a given test dataset, we randomly divide 5\% of the training
dataset into test sets for subsequent analysis and observations. We start from the two SFTed models and fine-tune them with these two datasets. As for the training parameter, we mainly follow the experimental detail in \cite{meng2024simpo}, and train the model for 1 epoch with a learning rate of 6e-7. The optimizer we choose is \texttt{paged\_Adamw\_32bit} with the implementation from \href{https://github.com/bitsandbytes-foundation/bitsandbytes/tree/main}{\texttt{bitsandbytes}}.

\textbf{Computation Environment.} The training experiments in this paper were conducted using 8×A800 80GB GPUs and 4×A100 40GB GPUs, based on the alignment handbook repository \cite{Tunstall_The_Alignment_Handbook}. Specifically, all Llama 3 8B-related experiments were performed on 4×A100 GPUs, while Qwen 2-7B experiments utilized 8×A800 GPUs. For downstream performance evaluation, MT-Bench was assessed using the \texttt{llm\_judge} submodule from the FastChat \cite{zheng2023judging} repository and the win rate experiments were conducted with the \href{https://github.com/ollama/ollama?tab=readme-ov-file}{\texttt{Ollama}} repository.

\subsection{Llama 3.3-70B Prompts for Win Rate Experiment}
\label{app:prompts}
\begin{verbatim}
You are tasked with comparing the responses of two assistants, Assistant A 
and Assistant B, to a user’s question. Additionally, you will be provided 
with a reference answer to evaluate the quality of the responses from both 
assistants.

First, output only 'A' or 'B' to indicate your judgment on which response is 
better. Then, provide a one-sentence explanation for your choice. The 
principles for your judgment should be based on the following criteria:

1. The most important criterion is to select the response whose meaning is 
essentially closer to the reference answer.
2. Do not judge the quality of the two responses based on their length.
3. Evaluate the responses based on their helpfulness, relevance, accuracy, 
depth, and conciseness.

User’s Question:
{question}

Reference Answer:
{ref_answer}

Assistant A’s Response:
{response_compare}

Assistant B’s Response:
{response_baseline}
\end{verbatim}

\clearpage
\subsection{Ablation Studies for \texttt{fixed}}
\label{app:candr_supp}

\begin{figure}[H]
    \centering
    \includegraphics[width=\linewidth]{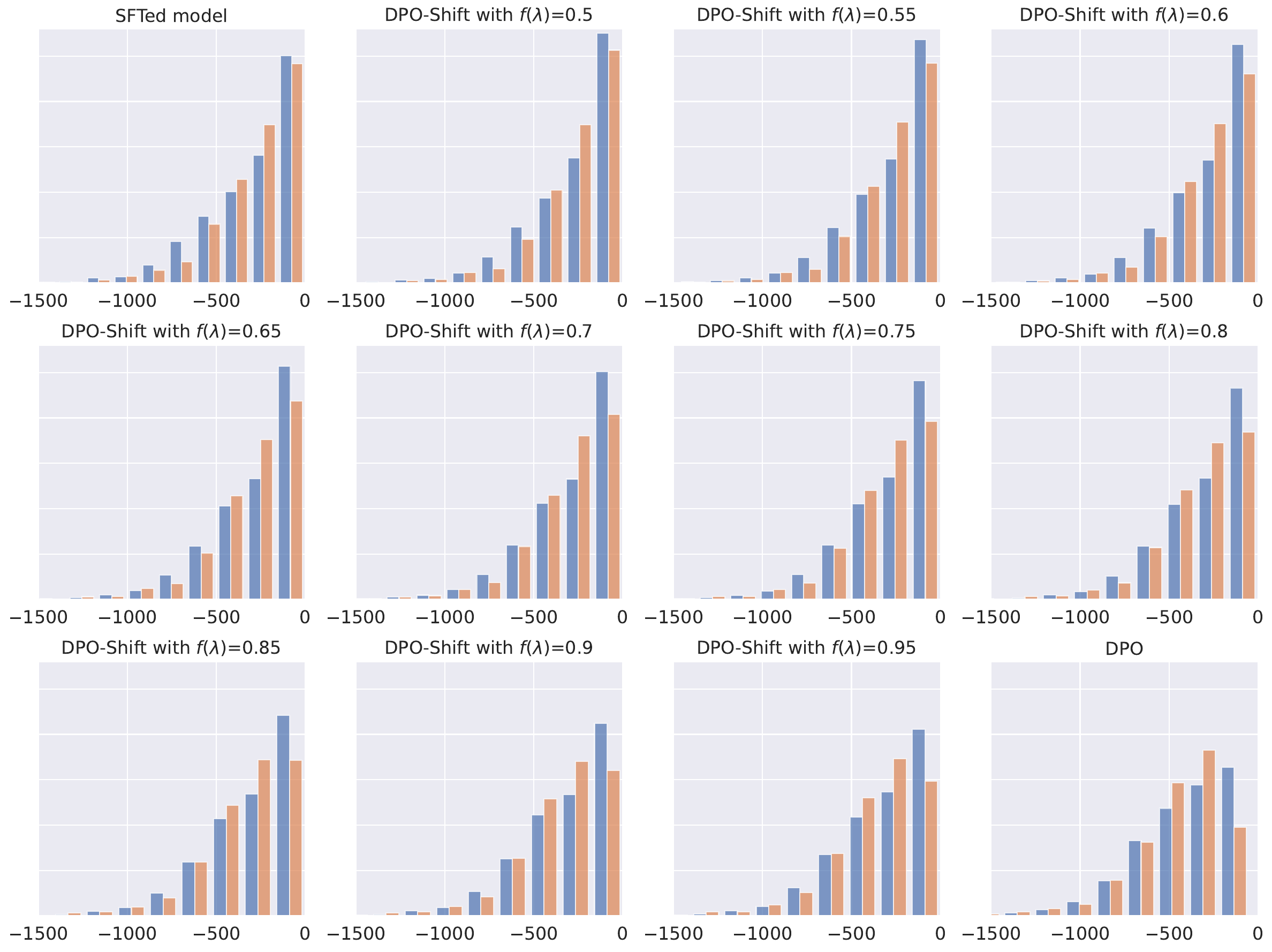}
    \caption{Distribution for $\logwin$ and $\logrej$ on test set split of UltraFeedback for Llama 3-8B trained on UltraFeedback, where \method uses \texttt{fixed} strategy. The ranges of the y-axis of all subfigures are the same. \label{fig:fixed-llama3-ultra-full}}
\end{figure}

\begin{figure}[H]
    \centering
    \includegraphics[width=\linewidth]{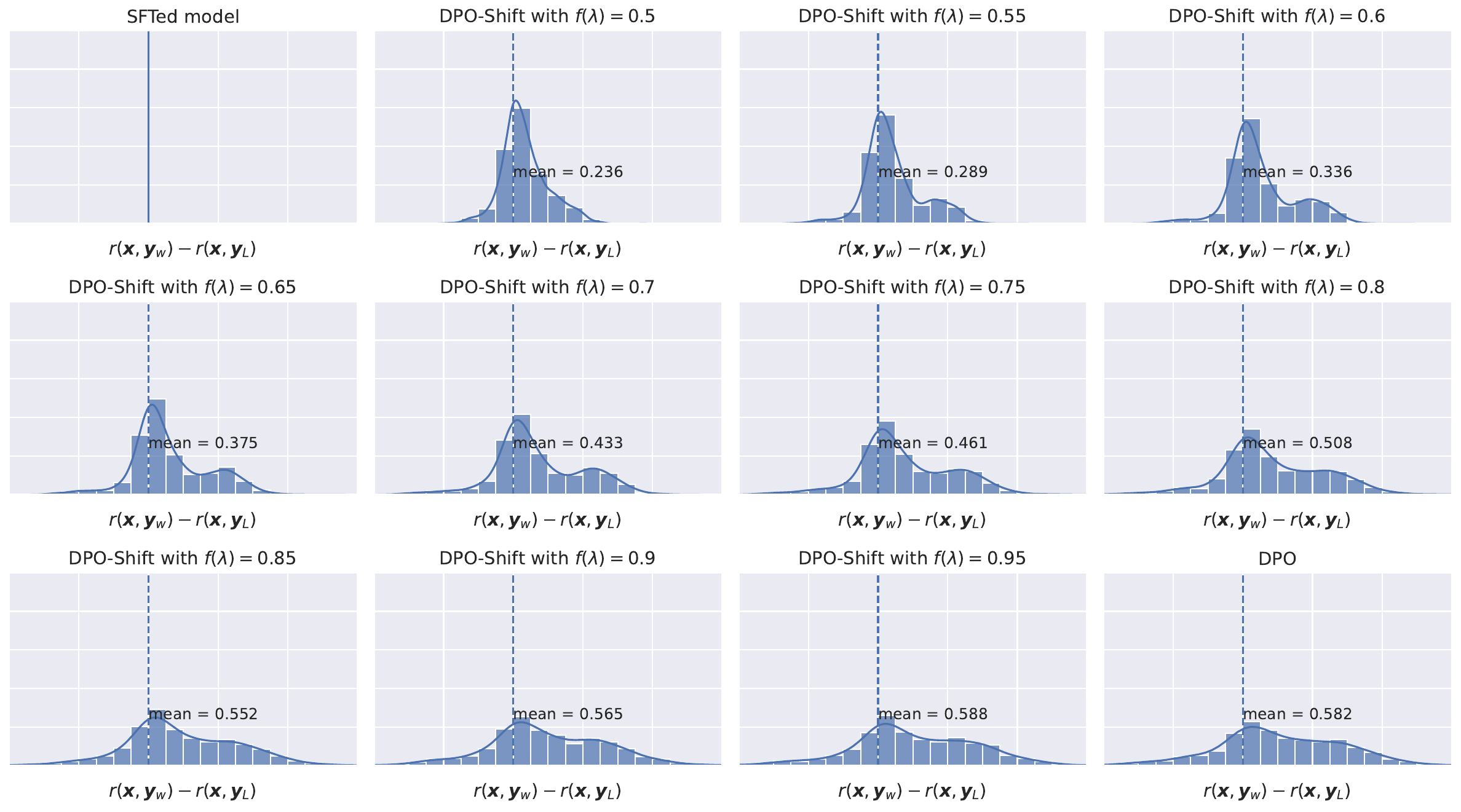}
    \caption{Distribution for reward margin and its mean on test set split of UltraFeedback for Llama 3-8B trained on UltraFeedback, where \method uses \texttt{fixed} strategy. The ranges of the y-axis of all subfigures are the same.}
\end{figure}

\clearpage
\begin{figure}[H]
    \centering
    \includegraphics[width=\linewidth]{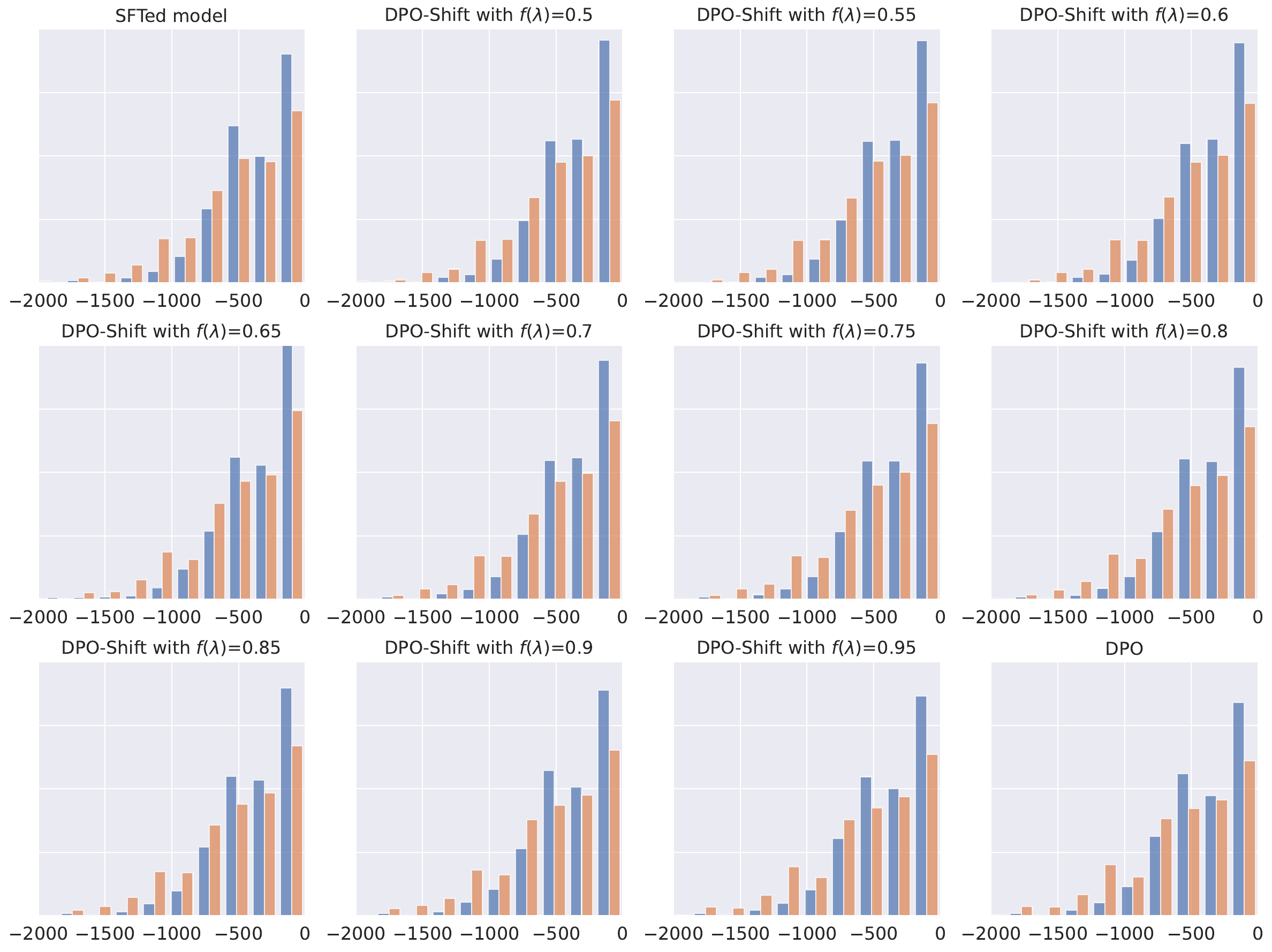}
    \caption{Distribution for $\logwin$ and $\logrej$ on test set split of Capybara for Llama 3-8B trained on Capybara, where \method uses \texttt{fixed} strategy. The ranges of the y-axis of all subfigures are the same.}
\end{figure}

\begin{figure}[H]
    \centering
    \includegraphics[width=\linewidth]{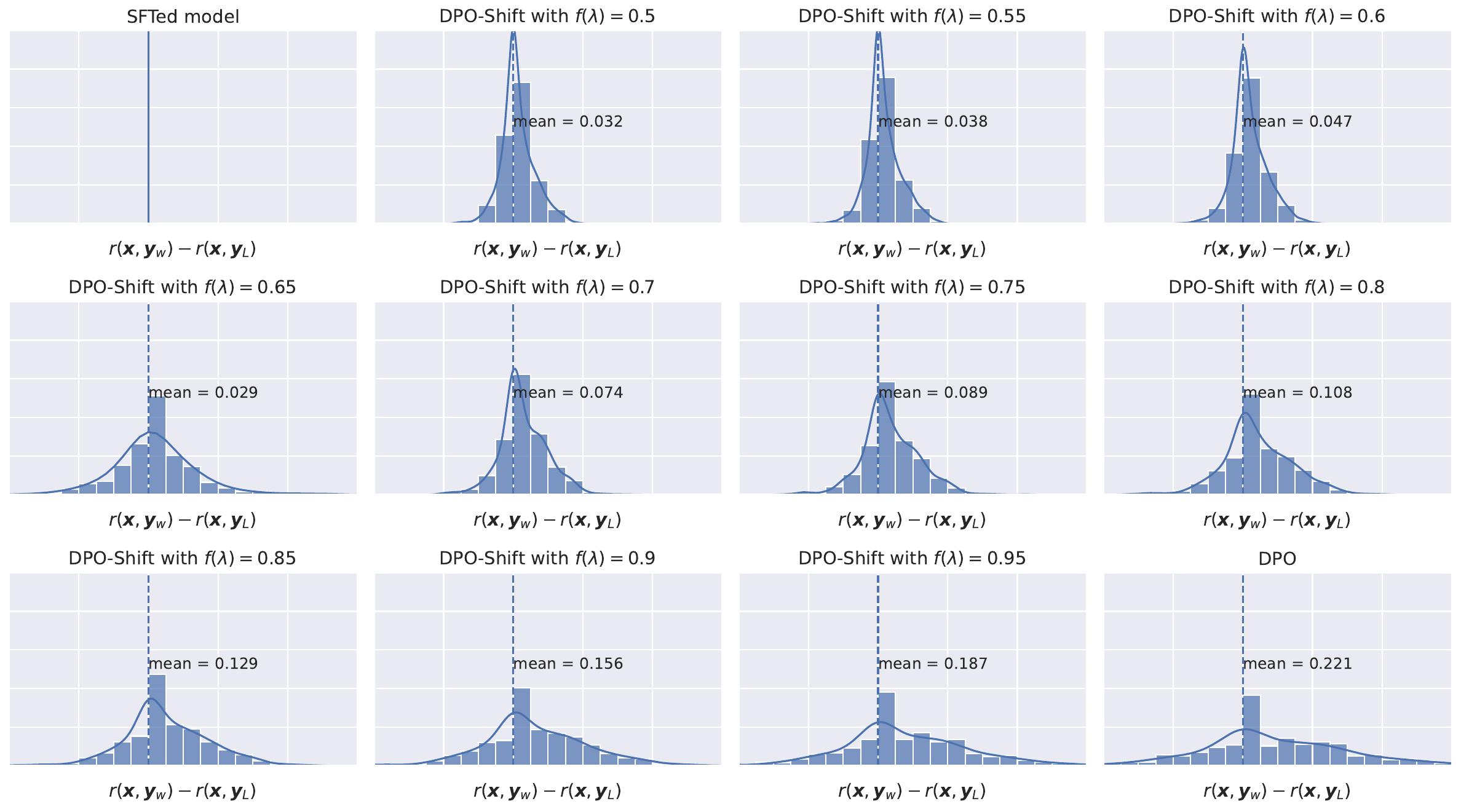}
    \caption{Distribution for reward margin and its mean on test set split of Capybara for Llama 3-8B trained on Capybara, where \method uses \texttt{fixed} strategy. The ranges of the y-axis of all subfigures are the same.}
\end{figure}

\clearpage

\begin{figure}[H]
    \centering
    \includegraphics[width=\linewidth]{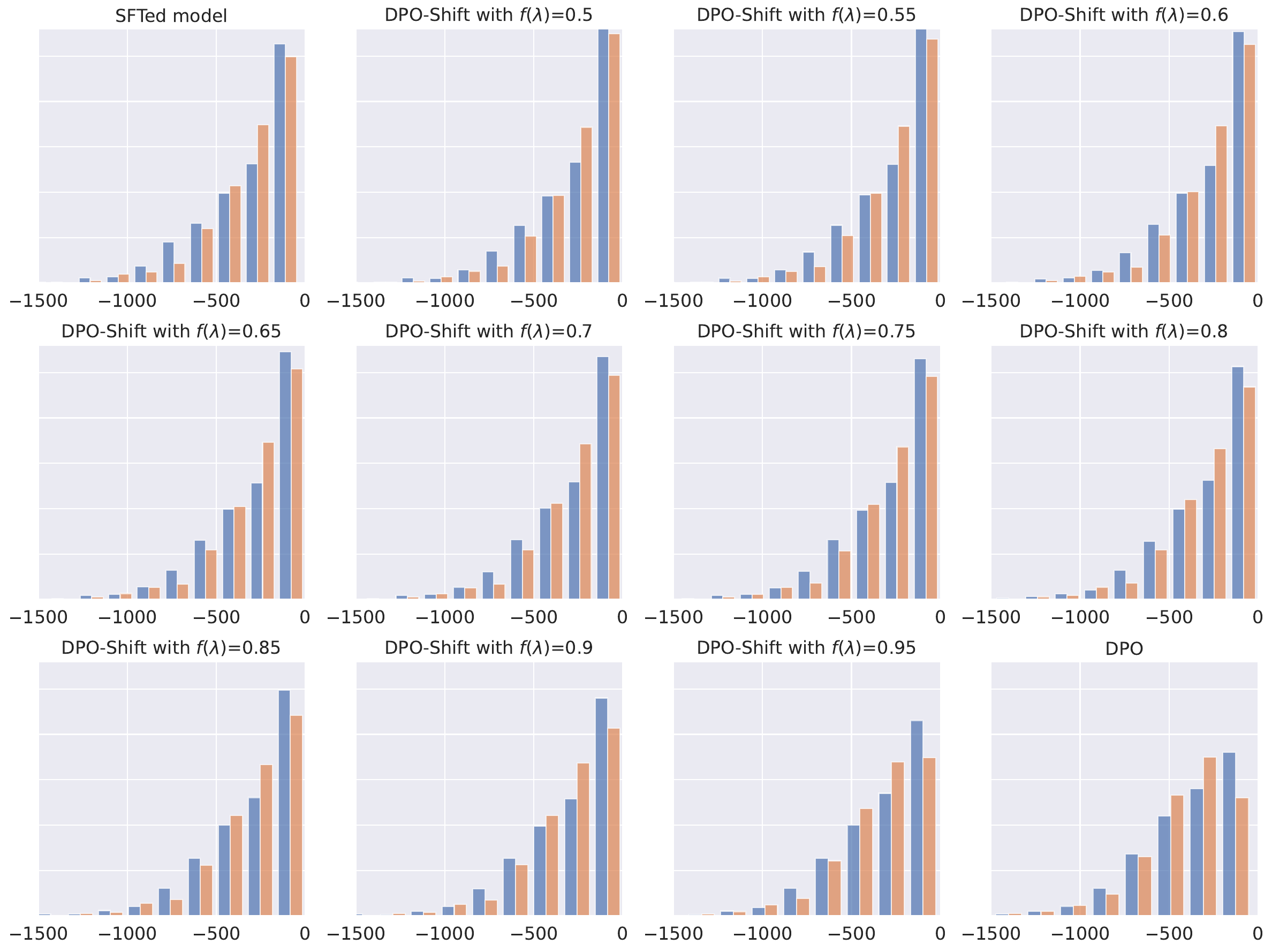}

    \caption{Distribution for $\logwin$ and $\logrej$ on test set split of UltraFeedback for Qwen 2-7B trained on UltraFeedback, where \method uses \texttt{fixed} strategy. The ranges of the y-axis of all subfigures are the same.}
\end{figure}

\begin{figure}[H]
    \centering
    \includegraphics[width=\linewidth]{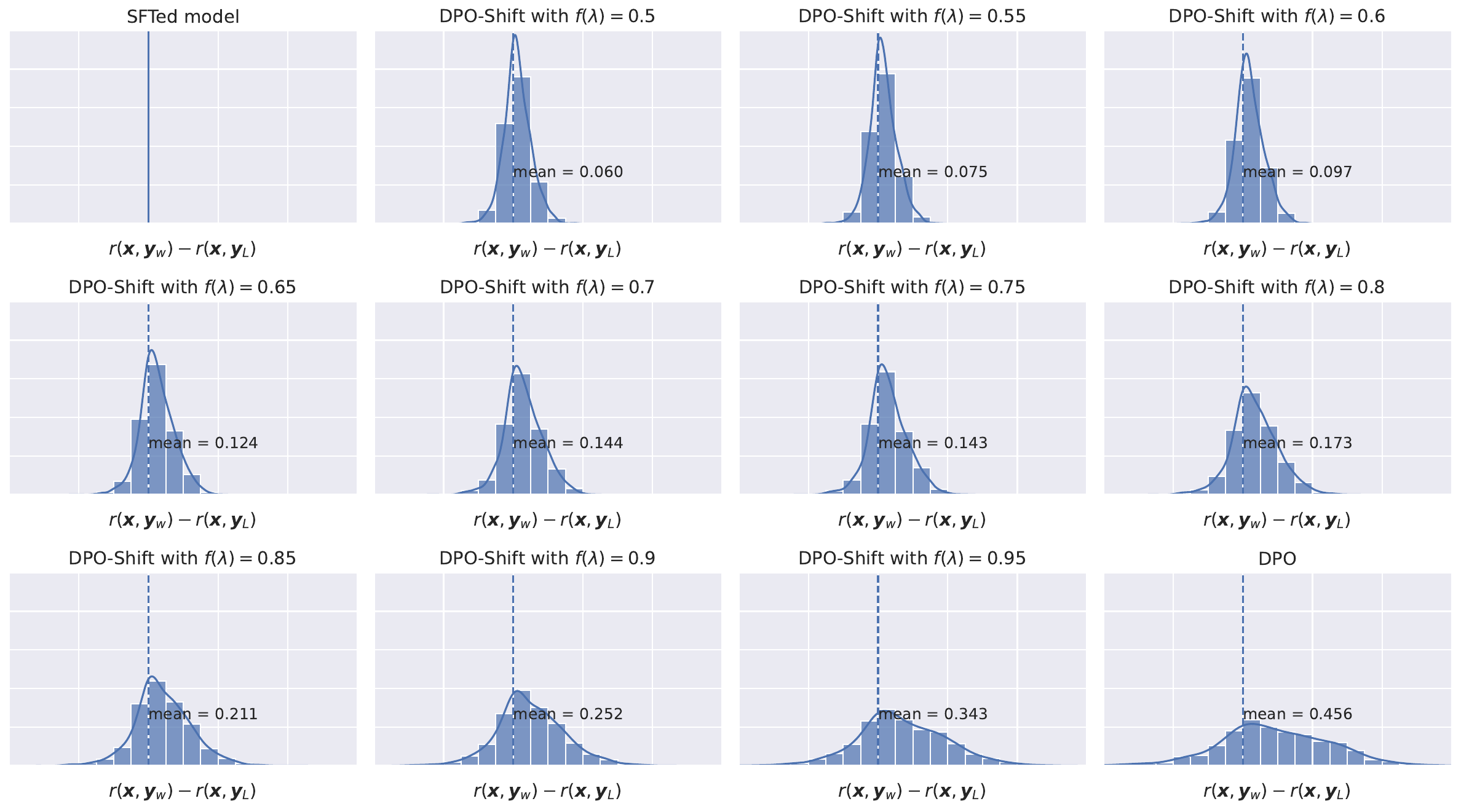}

    \caption{Distribution for reward margin and its mean on test set split of UltraFeedback for Qwen 2-7B trained on UltraFeedback, where \method uses \texttt{fixed} strategy. The ranges of the y-axis of all subfigures are the same.}
\end{figure}

\clearpage

\begin{figure}[H]
    \centering
    \includegraphics[width=\linewidth]{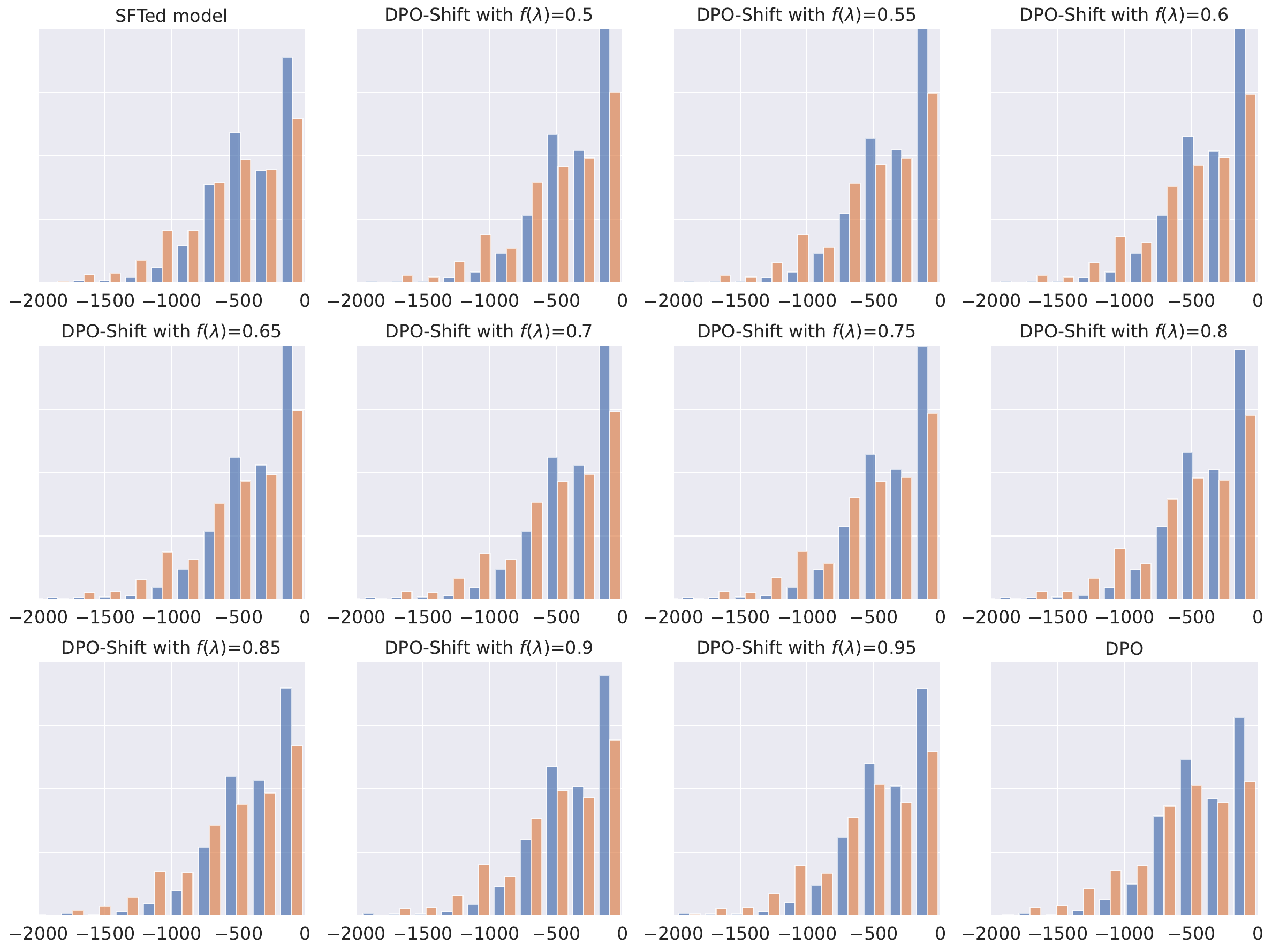}

    \caption{Distribution for $\logwin$ and $\logrej$ on test set split of Capybara for Qwen 2-7B trained on Capybara, where \method uses \texttt{fixed} strategy. The ranges of the y-axis of all subfigures are the same.}
\end{figure}

\begin{figure}[H]
    \centering
    \includegraphics[width=\linewidth]{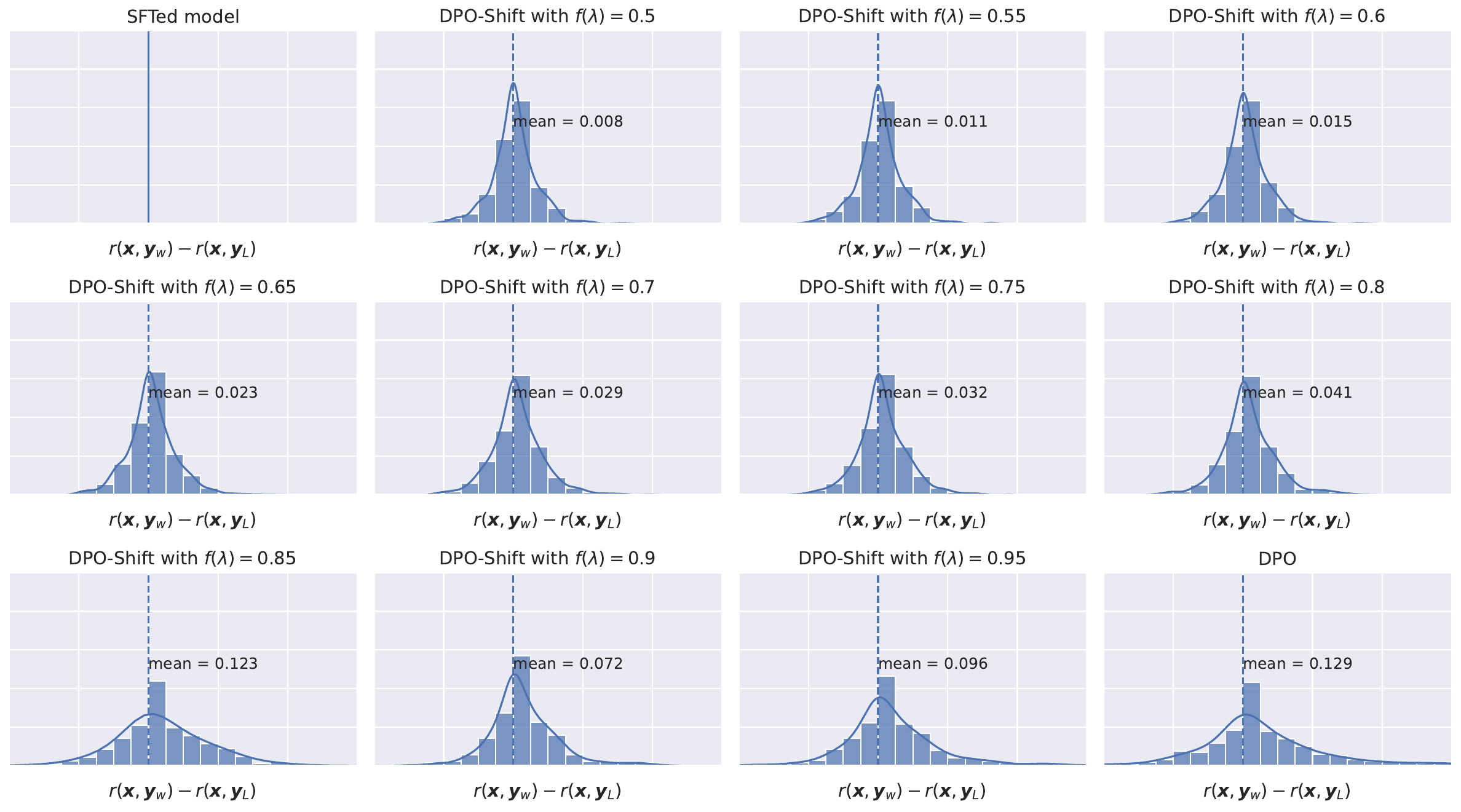}

    \caption{Distribution for reward margin and its mean on test set split of Capybara for Qwen 2-7B trained on Capybara, where \method uses \texttt{fixed} strategy. The ranges of the y-axis of all subfigures are the same.}
\end{figure}

\clearpage
\subsection{Ablation Studies for \texttt{linear\_increase} and \texttt{linear\_decrease}}
\label{app:candr_var_supp}

\begin{figure}[H]
    \centering
    \includegraphics[width=\linewidth]{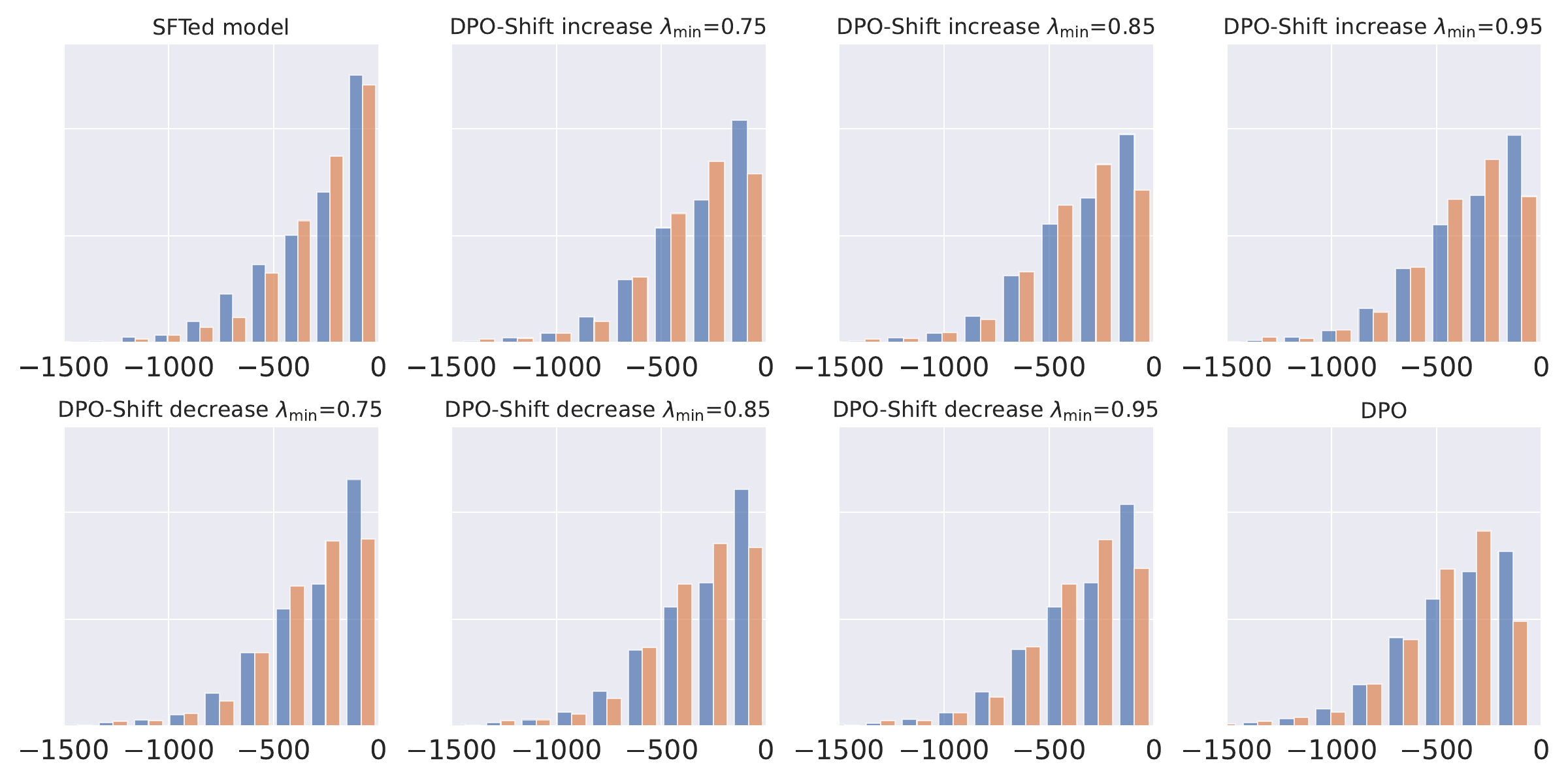}
    \vspace{-0.5cm}
    \caption{Distribution for $\logwin$ and $\logrej$ on test set split of Ultrafeedback for Llama 3-8B trained on UltraFeedback, where \method uses \texttt{linear\_increase} and \texttt{linear\_decrease} strategies. The ranges of the y-axis of all subfigures are the same.}
\end{figure}

\begin{figure}[H]
    \centering
    \includegraphics[width=\linewidth]{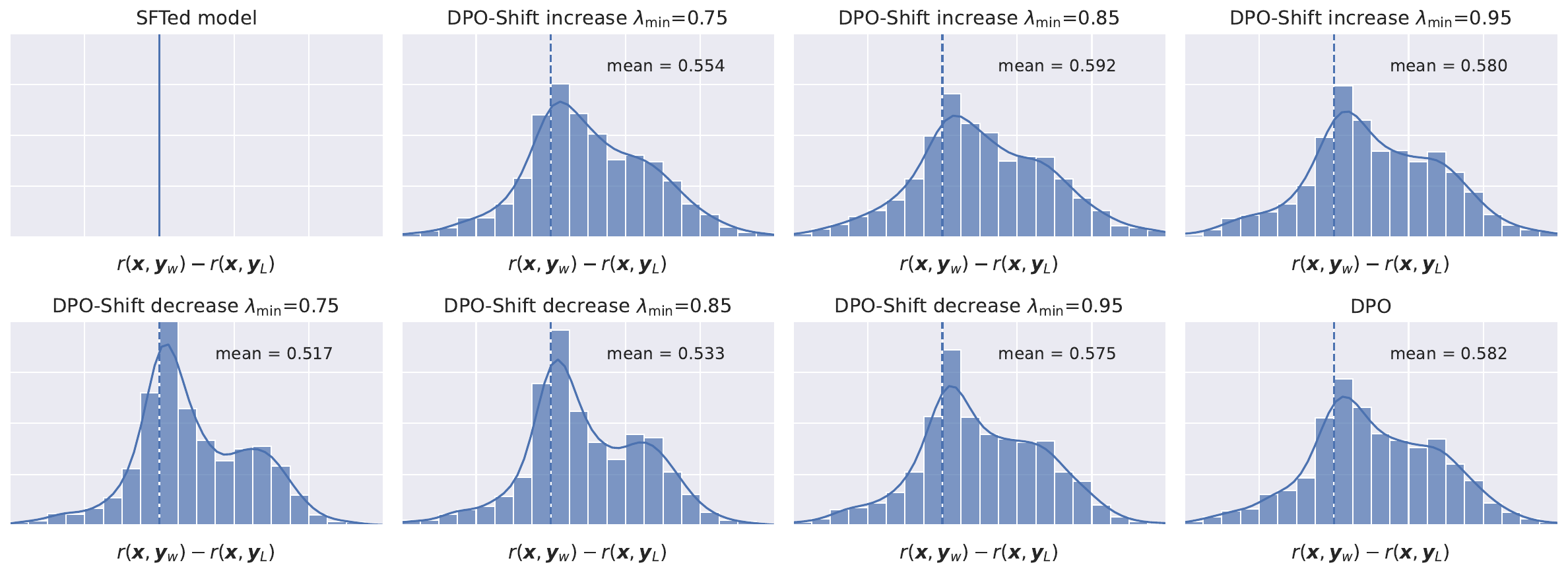}
    \caption{Distribution for reward margin and its mean on test set split of Ultrafeedback for Llama 3-8B trained on UltraFeedback, where \method uses \texttt{linear\_increase} and \texttt{linear\_decrease} strategies. The ranges of the y-axis of all subfigures are the same.}
\end{figure}

\clearpage
\begin{figure}[H]
    \centering
    \includegraphics[width=\linewidth]{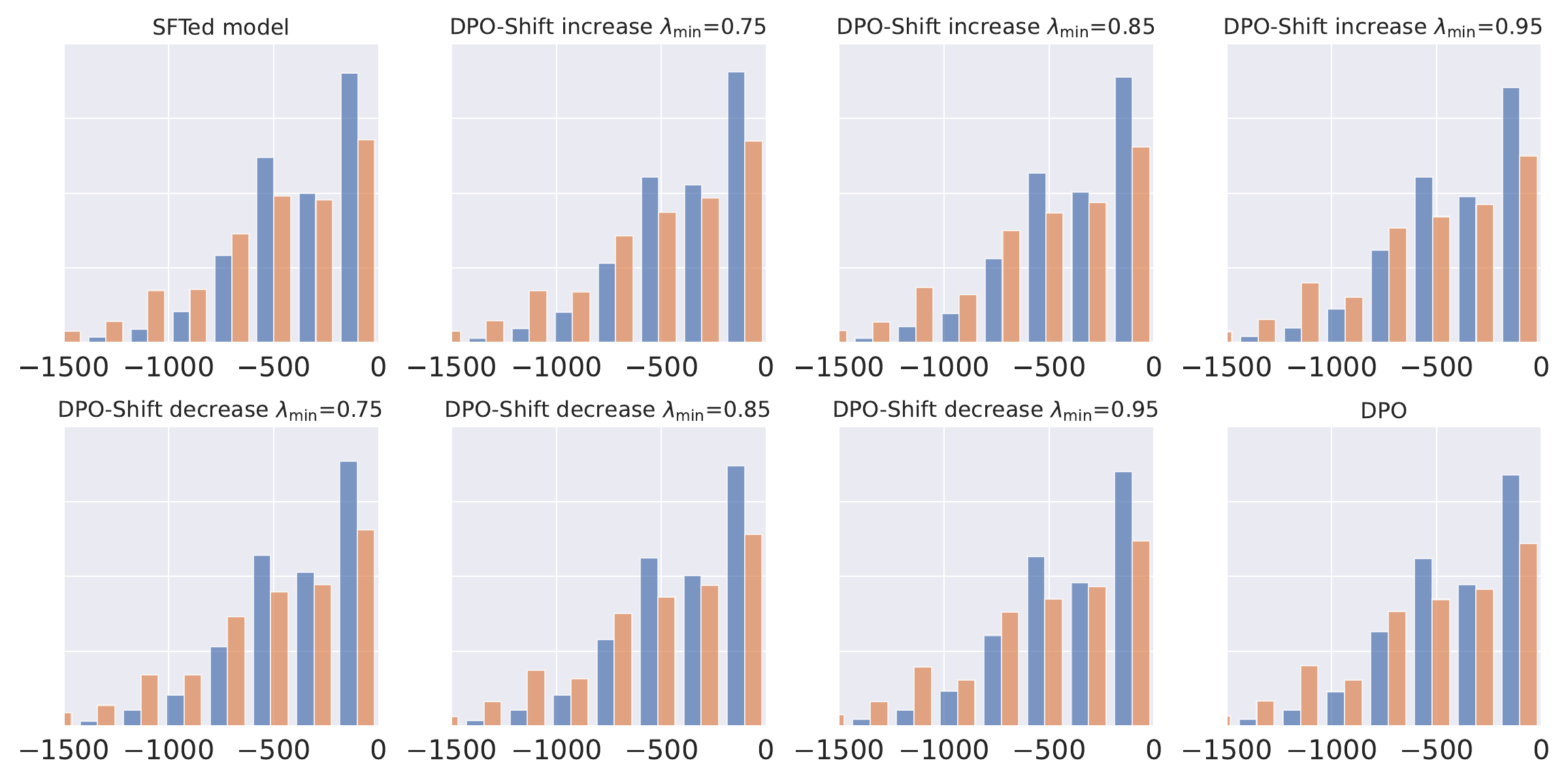}
    \caption{Distribution for $\logwin$ and $\logrej$ on test set split of Capybara for Llama 3-8B trained on Capybara, where \method uses \texttt{linear\_increase} and \texttt{linear\_decrease} strategies. The ranges of the y-axis of all subfigures are the same.}
\end{figure}

\begin{figure}[H]
    \centering
    \includegraphics[width=\linewidth]{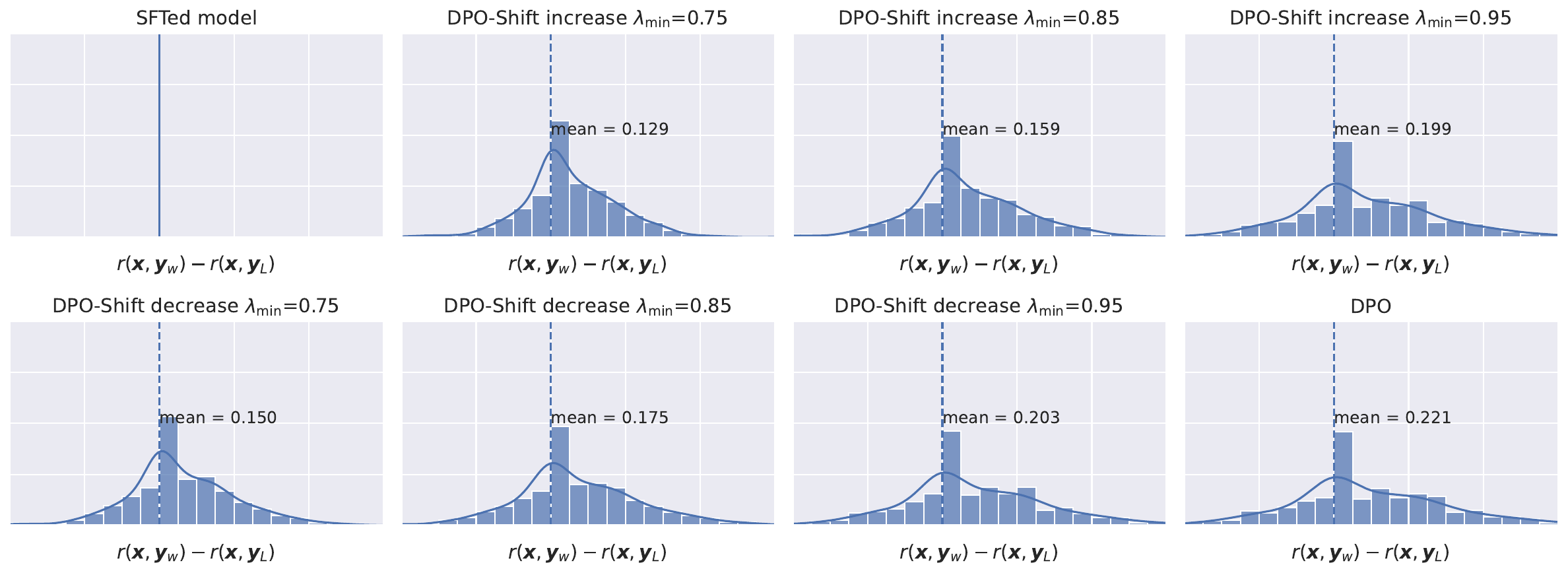}
    \caption{Distribution for reward margin and its mean on test set split of Capybara for Llama 3-8B trained on Capybara, where \method uses \texttt{linear\_increase} and \texttt{linear\_decrease} strategies. The ranges of the y-axis of all subfigures are the same.}
\end{figure}

\clearpage
\begin{figure}[H]
    \centering
    \includegraphics[width=\linewidth]{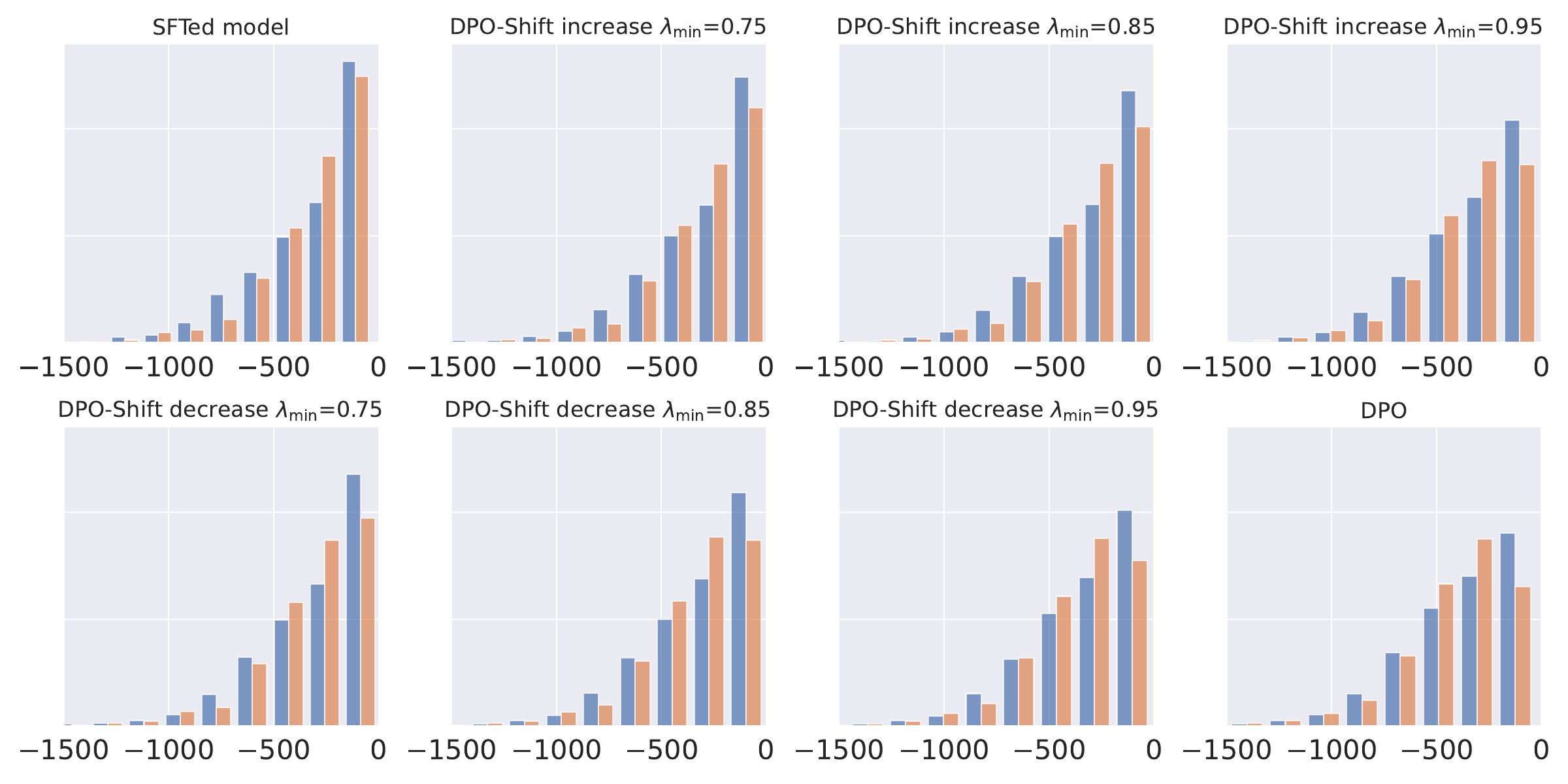}
    \caption{Distribution for $\logwin$ and $\logrej$ on test set split of UltraFeedback for Qwen 2-7B trained on UltraFeedback, where \method uses \texttt{linear\_increase} and \texttt{linear\_decrease} strategies. The ranges of the y-axis of all subfigures are the same.}
\end{figure}

\begin{figure}[H]
    \centering
    \includegraphics[width=\linewidth]{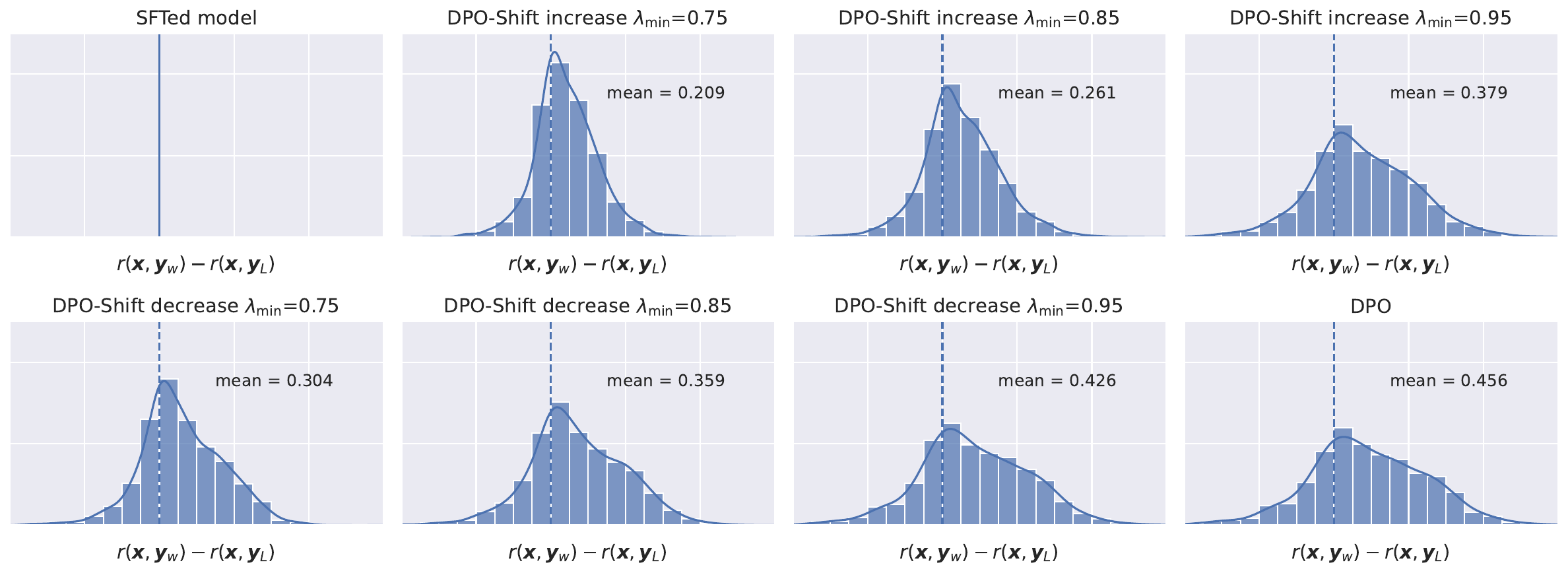}
    \caption{Distribution for reward margin and its mean on test set split of UltraFeedback for Qwen 2-7B trained on UltraFeedback, where \method uses \texttt{linear\_increase} and \texttt{linear\_decrease} strategies. The ranges of the y-axis of all subfigures are the same.}
\end{figure}

\clearpage
\begin{figure}[H]
    \centering
    \includegraphics[width=\linewidth]{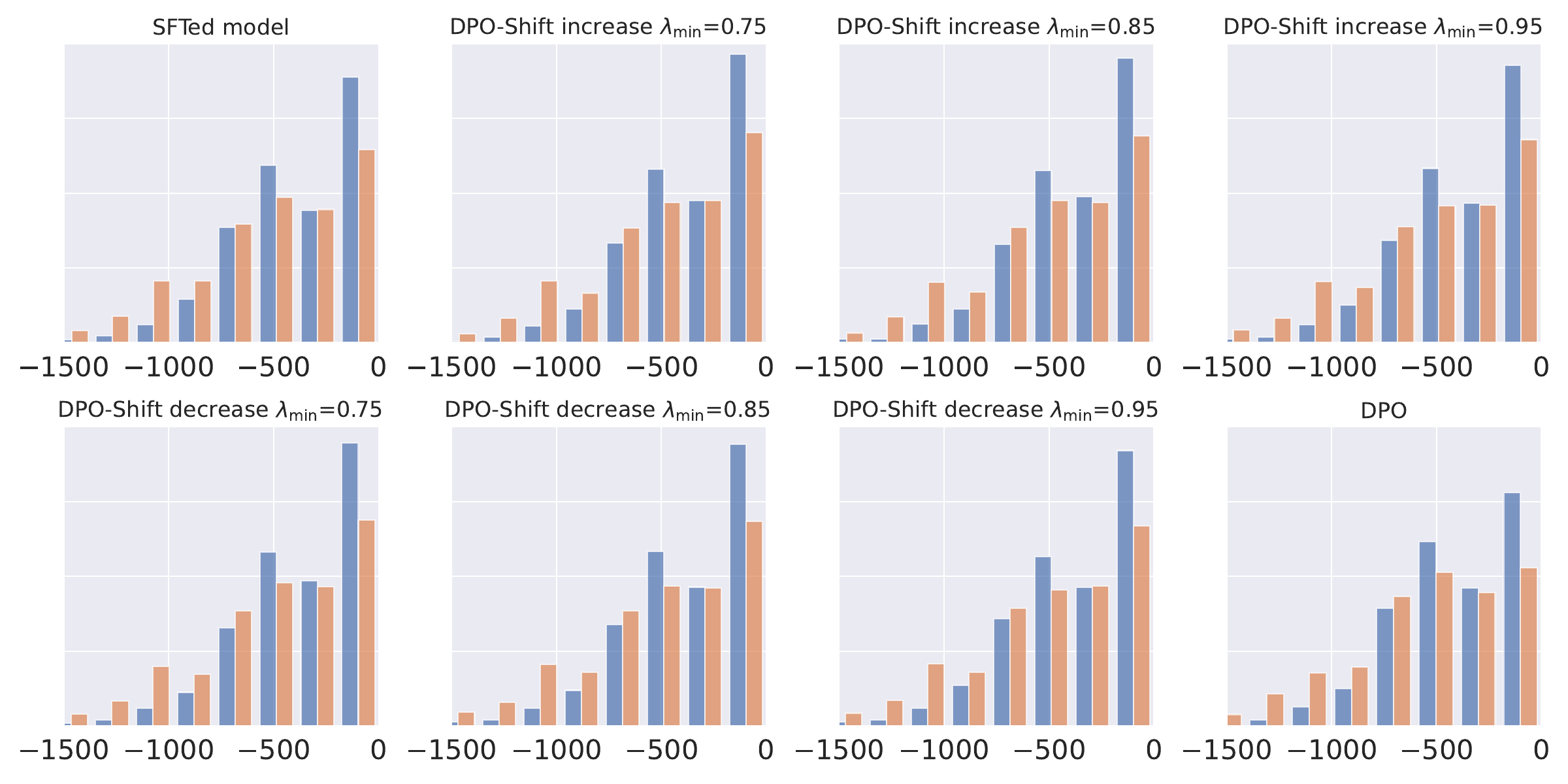}
    \caption{Distribution for $\logwin$ and $\logrej$ on test set split of Capybara for Qwen 2-7B trained on Capybara, where \method uses \texttt{linear\_increase} and \texttt{linear\_decrease} strategies. The ranges of the y-axis of all subfigures are the same.}
\end{figure}

\begin{figure}[H]
    \centering
    \includegraphics[width=\linewidth]{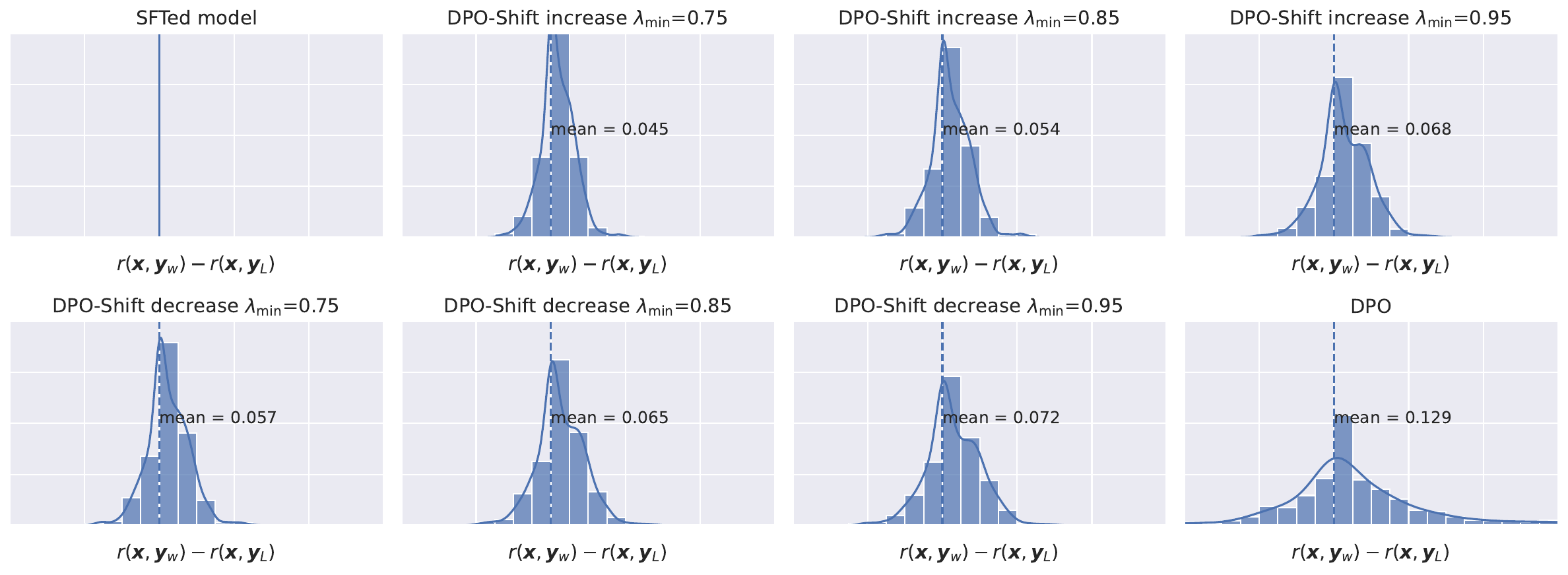}
    \caption{Distribution for reward margin and its mean on test set split of Capybara for Qwen 2-7B trained on Capybara, where \method uses \texttt{linear\_increase} and \texttt{linear\_decrease} strategies. The ranges of the y-axis of all subfigures are the same.}
\end{figure}

\clearpage
\subsection{Ablation Studies for Fine-grained \texttt{fixed}}
\label{app:phase_compare}
\begin{figure}[H]
    \centering
    \includegraphics[width=0.8\linewidth]{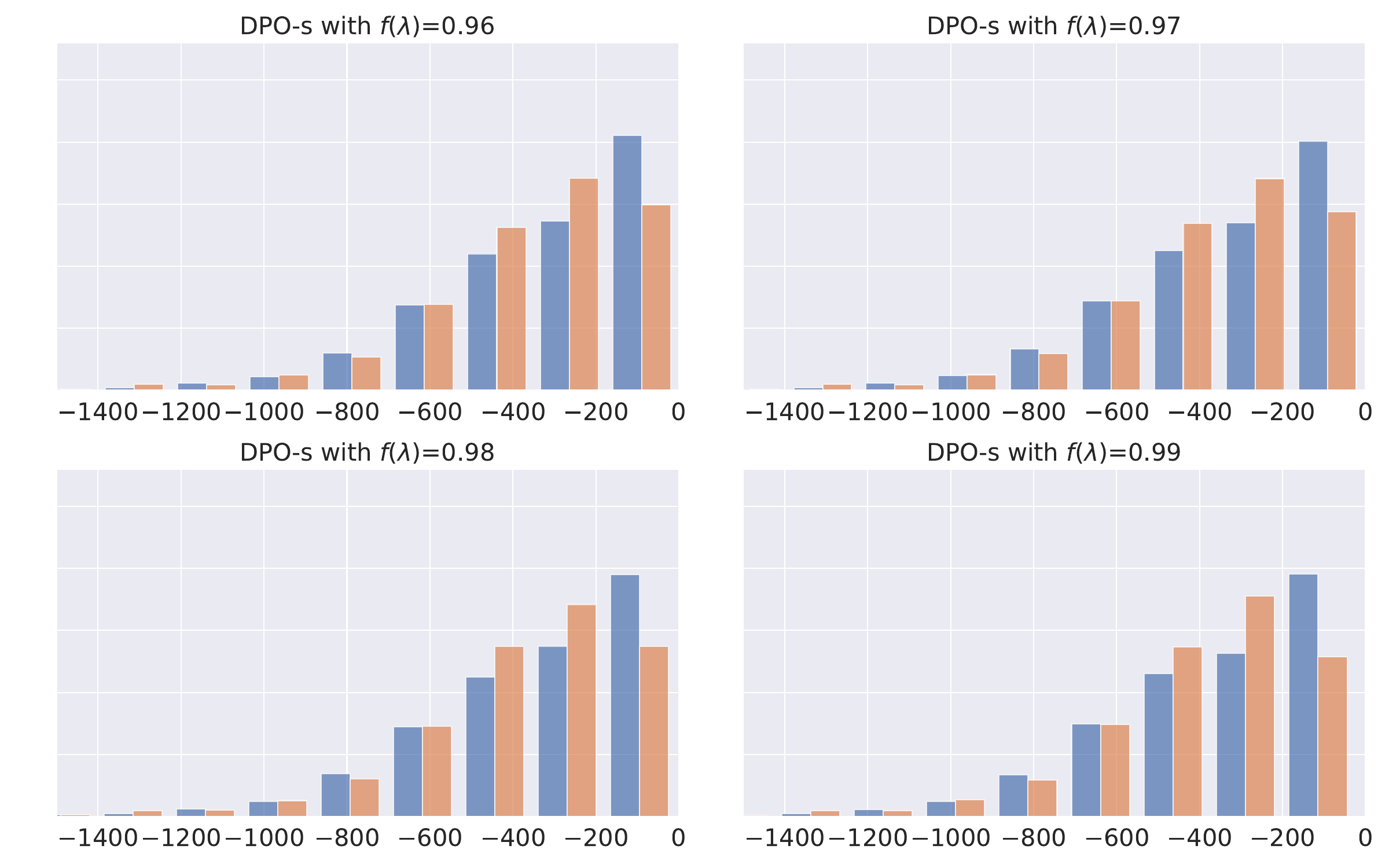}

    \caption{Distribution for $\logwin$ and $\logrej$ on test set split of UltraFeedback for Llama 3-8B trained on UltraFeedback, where \method uses \texttt{fixed} strategy. The ranges of the y-axis of all subfigures are the same.}
\end{figure}

\begin{figure}[H]
    \centering
    \includegraphics[width=0.8\linewidth]{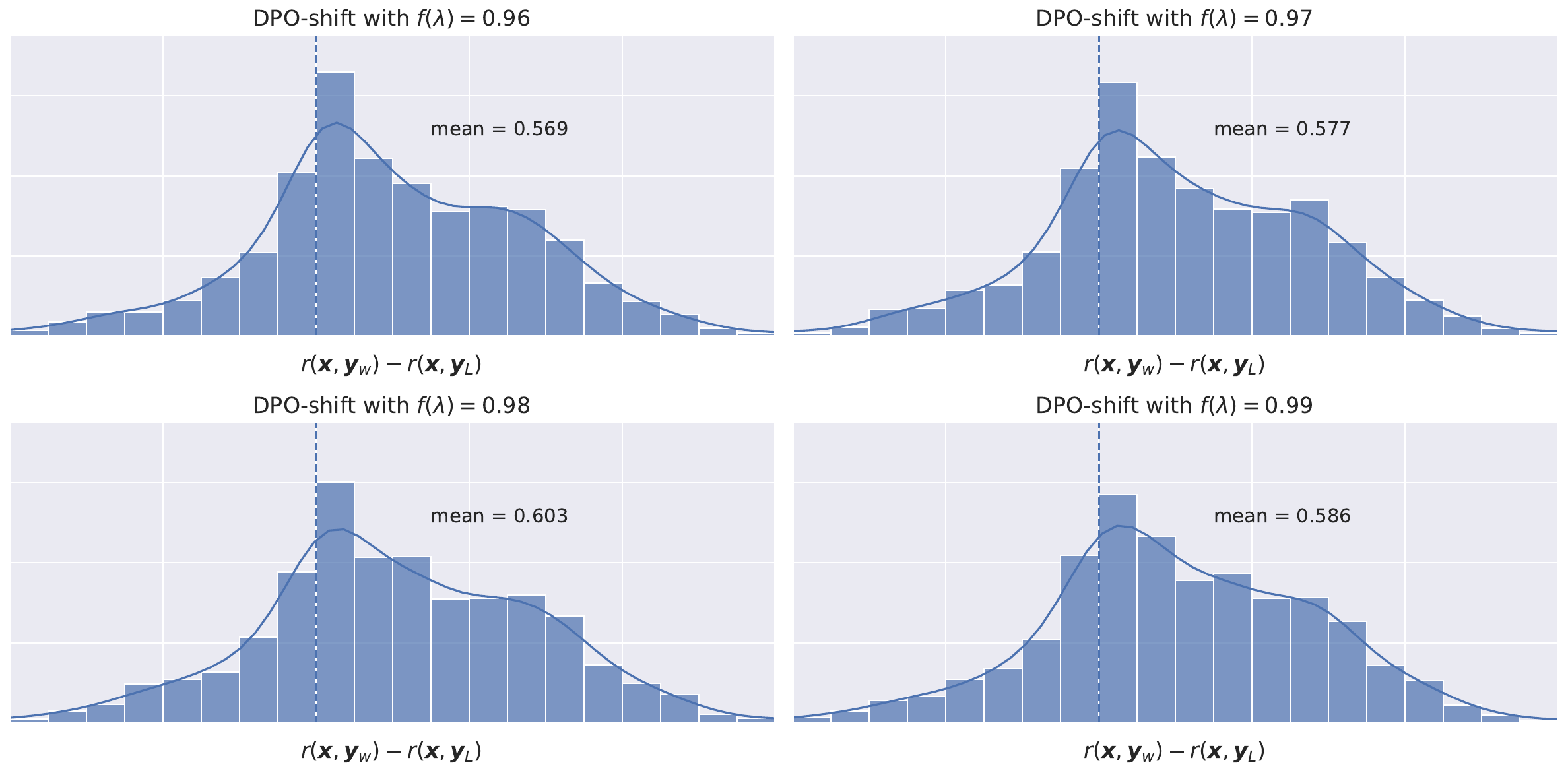}
    \caption{Distribution for reward margin and reward accuracy on test set split of UltraFeedback for Llama 3-8B trained on UltraFeedback, where \method uses \texttt{fixed} strategy. The ranges of the y-axis of all subfigures are the same.}
\end{figure}

\clearpage
\subsection{Supplementary Results for Win Rate Experiment}
\label{app:full: win rate}

\begin{table}[H]
\centering
\begin{tabular}{lcccc}
\toprule
$f(\lambda)$ strategy &  \textbf{Win} & \textbf{Lose} \\
\midrule
\textbf{SFT} & 30.70\% & 69.30\% \\
\midrule
\texttt{fixed} 0.55 & 49.15\% & 50.85\%  \\
\texttt{fixed} 0.75 & 59.40\% & 40.60\%  \\
\texttt{fixed} 0.95 & 69.60\% & 31.40\%  \\
\texttt{increase\_linear} 0.95 & 71.95\% & 29.05\% \\
\texttt{decrease\_linear} 0.95 & 73.30\% & 26.70\% \\
\bottomrule
\end{tabular}
\vspace{0.2cm}
\caption{Win rate experiment against DPO using Qwen-2 7B trained on the UltraFeedback dataset and tested with questions from the test split of UltraFeedback. Results for \method using \texttt{fixed}, \texttt{linear\_increase}, \texttt{linear\_decrease} are included.}
\label{table:win_rate_qwen}
\end{table}

\clearpage
\section{Proof of \Cref{thm:gap}}
\label{app:proof}

We consider our modified PO loss function:

\[
    \mathcal{L}_{\lambda-\text{DPO}}(\pi_{\theta},\pi_{\text{ref}})= -\mathbb{E}
    _{(\boldsymbol{x},\boldsymbol{y}_w,\boldsymbol{y}_l)\sim \mathcal{D}_{\text{pref}}}
    \left[\sigma\left(\beta\log\frac{\pi_{\theta}(\boldsymbol{y}_{w}|\boldsymbol{x})}{\pi_{\mathrm{ref}}(\boldsymbol{y}_{w}|\boldsymbol{x})}
    -f(\lambda)\cdot\beta\log\frac{\pi_{\theta}(\boldsymbol{y}_{l}|\boldsymbol{x})}{\pi_{\mathrm{ref}}(\boldsymbol{y}_{l}|\boldsymbol{x})}
    \right)\right]
\]
where $f(\lambda)$ is a real valued function with $f(\lambda)<1$. We compute
its gradient w.r.t $\theta$:
\begin{align*}
    \nabla_{\theta}\mathcal{L}_{\lambda-\text{DPO}}(\pi_{\theta},\pi_{\text{ref}}) & =-\mathbb{E}_{(\boldsymbol{x},\boldsymbol{y}_w,\boldsymbol{y}_l)\sim \mathcal{D}_{\text{pref}}}\left[\frac{\sigma'(u)}{\sigma(u)}\nabla_{\theta}u\right]                                                                                                       \\
    u                                                                              & :=\beta\log\frac{\pi_{\theta}(\boldsymbol{y}_{w}|\boldsymbol{x})}{\pi_{\mathrm{ref}}(\boldsymbol{y}_{w}|\boldsymbol{x})}-f(\lambda)\cdot\beta\log\frac{\pi_{\theta}(\boldsymbol{y}_{l}|\boldsymbol{x})}{\pi_{\mathrm{ref}}(\boldsymbol{y}_{l}|\boldsymbol{x})}
\end{align*}

notice that $\sigma'(x)=\sigma(x)(1-\sigma(x)),1-\sigma(x)=\sigma(-x).$

Then we proceed with the final gradient
\begin{align}
    \nabla_{\theta}\mathcal{L}_{\lambda-\text{DPO}}(\pi_{\theta},\pi_{\text{ref}})=-\mathbb{E}_{(\boldsymbol{x},\boldsymbol{y}_w,\boldsymbol{y}_l)\sim \mathcal{D}_{\text{pref}}}\left[\beta\sigma\left(f(\lambda)\cdot\beta\log\frac{\pi_{\theta}(\boldsymbol{y}_{l}|\boldsymbol{x})}{\pi_{\mathrm{ref}}(\boldsymbol{y}_{l}|\boldsymbol{x})}-\beta\log\frac{\pi_{\theta}(\boldsymbol{y}_{w}|\boldsymbol{x})}{\pi_{\mathrm{ref}}(\boldsymbol{y}_{w}|\boldsymbol{x})}\right)\right. \\
    \times\left.\left[ \nabla_{\theta}\log\pi \left(\boldsymbol{y}_{w}|\boldsymbol{x}\right)-f(\lambda)\cdot\nabla_{\theta}\log\pi \left(\boldsymbol{y}_{l}|\boldsymbol{x}\right) \right] \right]
\end{align}
We then simplify it with the following notation
\begin{align*}
    c_{1}:=c_{\theta}(\lambda,\boldsymbol{y}_{w},\boldsymbol{y}_{l}) & =\beta\sigma\left(f(\lambda)\cdot\beta\log\frac{\pi_{\theta}(\boldsymbol{y}_{l}|\boldsymbol{x})}{\pi_{\mathrm{ref}}(\boldsymbol{y}_{l}|\boldsymbol{x})}-\beta\log\frac{\pi_{\theta}(\boldsymbol{y}_{w}|\boldsymbol{x})}{\pi_{\mathrm{ref}}(\boldsymbol{y}_{w}|\boldsymbol{x})}\right) \\
    c_{2}                                                           & =f(\lambda)c_{1}
\end{align*}
then we have
\[
    \nabla_{\theta}\mathcal{L}_{\lambda-\text{DPO}}(\pi_{\theta},\pi_{\text{ref}}
    )=-\mathbb{E}_{(\boldsymbol{x},\boldsymbol{y}_w,\boldsymbol{y}_l)\sim
    \mathcal{D}_{\text{pref}}}\left[c_{1}\nabla_{\theta}\log\pi_{\theta} \left(\boldsymbol
    {y}_{w}|\boldsymbol{x}\right)-c_{2}\nabla_{\theta}\log\pi_{\theta} \left(\boldsymbol
    {y}_{l}|\boldsymbol{x}\right)\right].
\]
Then we upgrade $\theta_{t+1}$ with the following:
\[
    \theta_{t+1}\leftarrow \theta_{t}+\eta\mathbb{E}_{(\boldsymbol{x},\boldsymbol{y}_w,\boldsymbol{y}_l)\sim
    \mathcal{D}_{\text{pref}}}\left[c_{1}\nabla_{\theta}\log\pi_{\theta_t} \left(\boldsymbol
    {y}_{w}|\boldsymbol{x}\right)-c_{2}\nabla_{\theta}\log\pi_{\theta_t} \left(\boldsymbol
    {y}_{l}|\boldsymbol{x}\right)\right].
\]
We first look into
\[
    w_1(\theta_t)=\log\pi_{\theta_t} \left(\boldsymbol{y}_{w}|\boldsymbol{x}\right)
\]
then
\begin{align}
  w_1(\theta_{t+1}) & =w_1(\theta_t)+\eta\left( c_{1}\nabla_{\theta}\log\pi_{\theta_t} \left(\boldsymbol{y}_{w}|\boldsymbol{x}\right)-c_{2}\nabla_{\theta}\log\pi_{\theta_{t}} \left(\boldsymbol{y}_{l}|\boldsymbol{x}\right) \right)^{\top}\left( \nabla_{\theta}\log\pi_{\theta_{t}}\left(\boldsymbol{y}_{w}|\boldsymbol{x}\right) \right)     \\
    & =w_{1}(\theta_{t})+\eta \left( c_{1}\left||\nabla_{\theta}\log\pi_{\theta_{t}} \left(\boldsymbol{y}_{w}|\boldsymbol{x}\right)\right||^{2}-c_{2}\nabla_{\theta}\log\pi_{\theta_{t}} \left(\boldsymbol{y}_{l}|\boldsymbol{x}\right)^{\top}\nabla_{\theta}\log\pi_{\theta_{t}} \left(\boldsymbol{y}_{w}|\boldsymbol{x}\right)\right)
\end{align}

then
\begin{align*}
    g_1(t+1)&= \eta(c_{1}-c_{2})\nabla_{\theta}\log\pi \left(\boldsymbol{y}_{l}|\boldsymbol{x}\right)^{\top}\nabla_{\theta}\log\pi \left(\boldsymbol{y}_{w}|\boldsymbol{x}\right)\\
\end{align*}
We compute $\nabla_{\theta}\log\pi \left(\boldsymbol{y}_{l}^i|\boldsymbol{x}_i\right)^{\top}\nabla_{\theta}\log\pi \left(\boldsymbol{y}^i_{w}|\boldsymbol{x}_i\right)$ for the SFTed Llama 3-8B model. In terms of frequency, 71.4\% of them turn out to be positive. Consequently, we can choose a $c_{2}$ as small as possible to increase the chosen probability.

However, choosing small $c_{2}$ can cause performance drop. To evaluate the performance of the model, we look into the reward margin:
\[
    \omega_{2}(\theta_t)  =\mathbb{E}\left[\mathbf{1}\left\{\log\frac{\pi_{\theta_t}(\boldsymbol{y}_{w}|\boldsymbol{x})}{\pi_{\mathrm{ref}}(\boldsymbol{y}_{w}|\boldsymbol{x})} -\log\frac{\pi_{\theta_t}(\boldsymbol{y}_{l}|\boldsymbol{x})}{\pi_{\mathrm{ref}}(\boldsymbol{y}_{l}|\boldsymbol{x})}>0\right\}\right]
\]
To analyze, we alternate it with its smoothed version:
\[
    \omega_{2}(\theta_t)=\mathbb{E}\left[\sigma\left(\gamma\log\frac{\pi_{\theta_t}(\boldsymbol{y}_{w}|\boldsymbol{x})}{\pi_{\mathrm{ref}}(\boldsymbol{y}_{w}|\boldsymbol{x})} -\gamma\log\frac{\pi_{\theta_t}(\boldsymbol{y}_{l}|\boldsymbol{x})}{\pi_{\mathrm{ref}}(\boldsymbol{y}_{l}|\boldsymbol{x})}\right)\right],
\]
we abuse notation and use $\beta \rightarrow +\infty$ as hyper parameter.

Then with first order Taylor's expansion for
$\left( \omega',\theta' \right)$ and $\left( \omega,\theta \right)$.

\textbf{Remark 1.} Given the fact that they are using the same expectation
$\mathbb{E}_{(\boldsymbol{x},\boldsymbol{y}_w,\boldsymbol{y}_l)\sim
\mathcal{D}_{\text{pref}}}\left[\cdot\right]$, we omit it for the sake of simplicity
2. Given the assumption that $\eta$ is small enough, we can ignore second
order.
 \begin{align*}
    \omega_{2}(\theta_{t+1}) & =\omega_{2}(\theta_{t})+\eta(\theta^{2}_{t+1}-\theta^{2}_{t})^{\top}\sigma\left(\log\frac{\pi_{\theta_{t}} \left(\boldsymbol{y}_{l}|\boldsymbol{x}\right)}{\pi_{\mathrm{ref}}{(\boldsymbol{y}_{l}|\boldsymbol{x})}}-\log \frac{\pi_{\theta_{t}} \left(\boldsymbol{y}_{w}|\boldsymbol{x}\right)}{\pi_{\mathrm{ref}}(\boldsymbol{y_{w}|x})}\right)\cdot \left( \nabla_{\theta}\log\pi_{\theta_{t}} \left(\boldsymbol{y}_{w}|\boldsymbol{x}\right)-\nabla_{\theta}\log\pi \left(\boldsymbol{y}_{l}|\boldsymbol{x}\right) \right)     \\
    & =\omega^{2}_{t}+\eta_{1}\left( c_{1}\nabla_{\theta}\log\pi_{\theta_{t}} \left(\boldsymbol{y}_{w}|\boldsymbol{x}\right)-c_{2}\nabla_{\theta}\log\pi_{\theta_{t}} \left(\boldsymbol{y}_{l}|\boldsymbol{x}\right) \right)^{\top}\left( \nabla_{\theta}\log\pi_{\theta_{t}} \left(\boldsymbol{y}_{w}|\boldsymbol{x}\right)-\nabla_{\theta}\log\pi_{\theta_{t}} \left(\boldsymbol{y}_{l}|\boldsymbol{x}\right) \right)                                                 \\
    & =\omega^{2}_{t}+\eta_{1}[c_{1}\left||\nabla_{\theta}\log\pi_{\theta_{t}} \left(\boldsymbol{y}_{w}|\boldsymbol{x}\right)\right||^{2}+c_{2}\left||\nabla_{\theta}\log\pi_{\theta_{t}} \left(\boldsymbol{y}_{l}|\boldsymbol{x}\right)\right||^{2}\\
    &-\left( c_{1}+c_{2}\right) \nabla_{\theta}\log\pi_{\theta_{t}}\left(\boldsymbol{y}_{l}|\boldsymbol{x}\right)^{\top}\nabla_{\theta}\log\pi_{\theta_{t}} \left(\boldsymbol{y}_{w}|\boldsymbol{x}\right)] \\
    & \eta_{1}:=\eta \sigma \left( \log\pi_{\theta_{t}} \left(\boldsymbol{y}_{w}|\boldsymbol{x}\right)-\log\pi_{\theta_{t}} \left(\boldsymbol{y}_{l}|\boldsymbol{x}\right) \right)
\end{align*}

then we have

\begin{align*}
g_2(t+1) &=\eta_{1}\left( c_{2}-c_{1}\right)\left||\nabla_{\theta}\log\pi \left(\boldsymbol{y}_{l}|\boldsymbol{x}\right)\right||^{2}+(c_{1}-c_{2})\nabla_{\theta}\log\pi \left(\boldsymbol{y}_{l}|\boldsymbol{x}\right)^{\top}\nabla_{\theta}\log\pi \left(\boldsymbol{y}_{w}|\boldsymbol{x}\right)   \\
& =\eta_{1}\left(c_{1}-c_{2}\right) \left(\nabla_{\theta}\log\pi \left(\boldsymbol{y}_{l}|\boldsymbol{x}\right)^{\top}\nabla_{\theta}\log\pi \left(\boldsymbol{y}_{w}|\boldsymbol{x}\right)-\left||\nabla_{\theta}\log\pi \left(\boldsymbol{y}_{l}|\boldsymbol{x}\right)\right||^{2}\right)
\end{align*}

We compute $\nabla_{\theta}\log\pi_{\theta_{t}} \left(\boldsymbol{y}_{l}^i|\boldsymbol{x}_i\right)^{\top}\nabla_{\theta}\log\pi_{\theta_{t}} \left(\boldsymbol{y}_{w}^{i}|\boldsymbol{x}_{i}\right)-\left||\nabla_{\theta}\log\pi_{\theta_{t}} \left(\boldsymbol{y}_{l}^{i}|\boldsymbol{x}_{i}\right)\right||^{2}$ for the SFTed Llama 3-8B model. In terms of frequency, 81.7\% of them turn out to be positive.

which is consistent with our observation that accuracy drops greatly with small
$c_{2}$ , therefore we need to strike a balance between the accuracy and chosen
probability.

\end{document}